\documentclass[10pt,twocolumn,letterpaper]{article}

\usepackage[pagenumbers]{cvpr} %

\usepackage[accsupp]{axessibility}
\usepackage[dvipsnames]{xcolor}
\usepackage{graphicx}
\usepackage[percent]{overpic}

\definecolor{color1}{HTML}{FFA500}
\definecolor{color2}{HTML}{0000FF}
\definecolor{color3}{HTML}{008000}

\definecolor{cvprblue}{rgb}{0.21,0.49,0.74}
\usepackage[pagebackref,breaklinks,colorlinks,citecolor=cvprblue]{hyperref}

\definecolor{darkpastelgreen}{rgb}{0.01, 0.75, 0.24}
\definecolor{darkgreen}{rgb}{0.00, 0.8, 0.2}
\definecolor{darkyellow}{rgb}{0.96, 0.75, 0.00}

\title{Producing and Leveraging Online Map Uncertainty in Trajectory Prediction}

\author{Xunjiang Gu$^{1}$ \hspace{0.6cm} Guanyu Song$^{1}$ \hspace{0.6cm} Igor Gilitschenski$^{1, 2}$ \hspace{0.6cm} Marco Pavone$^{3, 4}$ \hspace{0.6cm} Boris Ivanovic$^{3}$\\
$^{1}$University of Toronto \hspace{0.6cm} $^{2}$Vector Institute \hspace{0.6cm} $^{3}$NVIDIA Research \hspace{0.6cm} $^{4}$Stanford University\\
{\tt\small \{alfred.gu, guanyu.song\}@mail.utoronto.ca, gilitschenski@cs.toronto.edu,}\\
{\tt\small \{mpavone, bivanovic\}@nvidia.com, pavone@stanford.edu}
}

\begin{document}

\setlength{\abovedisplayskip}{3pt}
\setlength{\belowdisplayskip}{3pt}

\maketitle
\begin{abstract}
High-definition (HD) maps have played an integral role in the development of modern autonomous vehicle (AV) stacks, albeit with high associated labeling and maintenance costs. As a result, many recent works have proposed methods for estimating HD maps online from sensor data, enabling AVs to operate outside of previously-mapped regions. However, current online map estimation approaches are developed in isolation of their downstream tasks, complicating their integration in AV stacks. In particular, they do not produce uncertainty or confidence estimates. In this work, we extend multiple state-of-the-art online map estimation methods to additionally estimate uncertainty and show how this enables more tightly integrating online mapping with trajectory forecasting\footnote{Code: \href{https://github.com/alfredgu001324/MapUncertaintyPrediction}{https://github.com/alfredgu001324/MapUncertaintyPrediction}}. In doing so, we find that incorporating uncertainty yields up to 50\% faster training convergence and up to 15\% better prediction performance on the real-world nuScenes driving dataset.
\end{abstract}    
\vspace{-0.3cm}

\section{Introduction}
\label{sec:intro}
\vspace{-0.1cm}
A critical component of autonomous driving is understanding the static environment, e.g., road layout and connectivity, surrounding the autonomous vehicle (AV). Accordingly, high-definition (HD) maps have been developed to capture and provide such information, containing semantics like road boundaries, lane dividers, and road markings at the centimeter level.
In recent years, HD maps have proven to be indispensable for AV development and deployment, seeing widespread use today~\cite{WaymoSafety2021}. However, HD maps are costly to label and maintain over time, and they can only be used in geofenced areas, limiting AV scalability.

\begin{figure}[t]
\centering

\vspace{-0.9cm}

\includegraphics[width=0.31\linewidth]{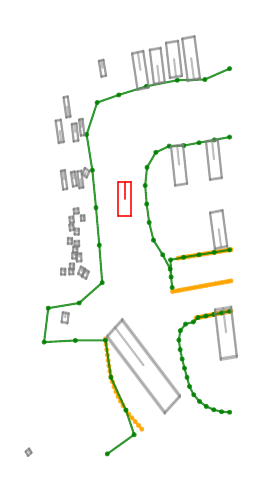}
\includegraphics[width=0.31\linewidth]{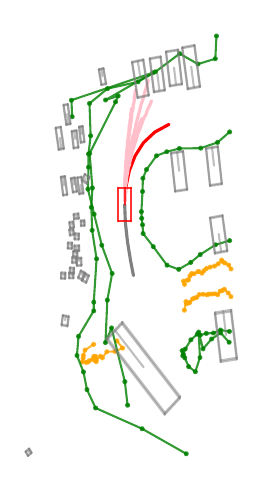}
\includegraphics[width=0.33\linewidth]{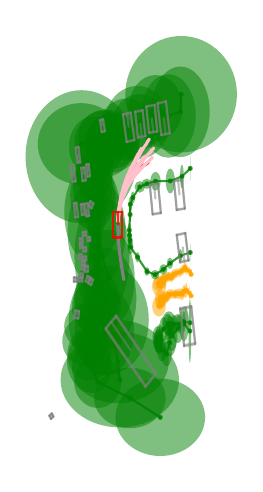}

\vspace{-0.6cm}

\caption{Producing uncertainty from online HD map estimation methods and incorporating it in downstream modules yields a variety of benefits.
\textbf{Left:} Ground truth HD map and agent positions. \textbf{Middle:} HiVT~\cite{zhou2022hivt} predictions using the map output by MapTR~\cite{MapTR}. \textbf{Right:} HiVT~\cite{zhou2022hivt} predictions using the map output by MapTR~\cite{MapTR} augmented with point uncertainties (which are large as the left road boundary is occluded by parked vehicles).}
\label{fig:hero}

\vspace{-0.6cm}

\end{figure}

To address these issues, many recent works turn to estimating HD maps online from sensor data. Broadly, they aim to predict the locations and classes of map elements, typically as polygons or polylines, all from camera images and LiDAR scans. However, current online map estimation methods do not produce any associated uncertainty or confidence information. This is problematic as it causes downstream consumers to implicitly assume that inferred map components are certain, and any mapping errors (e.g., shifting or incorrectly-placed map elements) may yield errant downstream behaviors. Towards this end, we propose to expose map uncertainty from online map estimation approaches and incorporate it in downstream modules. Concretely, we incorporate map uncertainty into trajectory prediction and find significant performance improvements in combined mapper-predictor systems with map uncertainty (\cref{fig:hero}) compared to those without.

\textbf{Contributions.} Our core contributions are threefold:
First, we propose a general vectorized map uncertainty formulation and extend multiple state-of-the-art online map estimation methods to additionally output uncertainty estimates, without any degradation in pure mapping performance.
Second, we empirically analyze potential sources of map uncertainty, confirming where current map estimation methods lack confidence and informing future research directions.
Third, we combine many recent online map estimation models with multiple state-of-the-art trajectory prediction approaches and show how incorporating online mapping uncertainty significantly improves the performance and training characteristics of downstream prediction models, speeding up training convergence by up to \textbf{50}\% and improving online prediction accuracy by up to \textbf{15}\%.

\section{Related Work}
\label{sec:related_work}

\subsection{Online Map Estimation}

The goal of online map estimation is to predict a representation of the static world elements surrounding an autonomous vehicle from sensor data. Initial works focused on producing 2D birds-eye-view (BEV) rasterized semantic segmentations as world representations by unprojecting to 3D and collapsing along the $Z$-axis~\cite{philion2020lss,liu2022bevfusion} or by leveraging cross-attention in geometry-aware Transformer~\cite{VaswaniShazeerEtAl2017} models~\cite{can2021stsu,li2022bevformer}.

Recently, vectorized map estimation approaches have emerged, extending BEV rasterization approaches with decoders that regress and classify polyline and polygon map elements (among other curve representations~\cite{qiao2023bemapnet}). Initial works such as HDMapNet~\cite{li2022hdmapnet} and SuperFusion~\cite{dong2022SuperFusion} propose to fuse LiDAR point clouds and RGB images into a common BEV feature frame followed by a hand-crafted post-processing step to produce polyline map elements. To remove the reliance on hand-crafted post-processing, VectorMapNet~\cite{liu2022vectormapnet} and InstaGraM~\cite{shin2023instagram} introduce end-to-end models for vectorized HD map learning. Further improvements to avoid information loss from key-point sampling along polylines are proposed by PivotNet~\cite{ding2023pivotnet}.

The MapTR line of work~\cite{MapTR, maptrv2} and extensions~\cite{xu2023insightmapper} formulate vectorized HD map estimation as a point set prediction task, yielding significant improvements in mapping performance.
Most recently, StreamMapNet~\cite{yuan2024streammapnet} focuses on incorporating temporal data from past frames, enabling HD map estimation from streaming data online.
In each of these methods, however, there is no uncertainty or confidence information provided to downstream consumers, making it difficult to distinguish between accurate and errant map elements.

\subsection{Map-Informed Trajectory Prediction}

Early trajectory prediction works predominantly leveraged rasterized maps to represent and encode scene context~\cite{RudenkoPalmieriEtAl2019}. Typically, a Convolutional Neural Network (CNN) encodes the BEV map tensor into a vector which is concatenated with other scene context (e.g., agent state histories) and passed through the rest of the model~\cite{SalzmannIvanovicEtAl2020,Phan-MinhGrigoreEtAl2020,YuanWengEtAl2021,GillesSabatiniEtAl2021,IvanovicHarrisonEtAl2023}.

Recently, trajectory prediction works have increasingly turned to encoding raw polyline information from vectorized HD maps, achieving significant performance improvements. Initial approaches~\cite{gao2020vectornet,liang2020lanegcn,ZhaoGaoEtAl2020,GillesSabatiniEtAl2022a,GillesSabatiniEtAl2022b} applied Graph Neural Networks (GNNs) to encode lane polylines and their influence on agent motion. Extending this idea, most current approaches adopt Transformer~\cite{VaswaniShazeerEtAl2017} architectures with map-agent cross-attention~\cite{liu2021mmtransformer,GuSunEtAl2021,zhou2022hivt,deo2023pgp} to achieve state-of-the-art performance.

In contrast to rasterized approaches, directly encoding vectorized HD maps removes information bottlenecks (discretization in rasterization loses fine geometric details) and enables a direct focus on map elements that are most relevant to agents (as opposed to encoding an entire BEV map of the scene). One core drawback, however, is a lack of uncertainty representations. While uncertainty can be naturally encoded in a BEV format (e.g., as a probability heatmap), how to best represent it in vectorized HD maps remains an open question. Our work addresses this problem by proposing a simple yet general methodology to estimate vectorized HD map uncertainty and represent it. 

\subsection{End-to-End Driving Architectures}

End-to-end AV architectures are a promising approach to developing integrated stacks which account for mapping uncertainty. Recently, UniAD~\cite{hu2023uniad}, VAD~\cite{jiang2023vad}, and OccNet~\cite{tong2023occnet} have shown how to incorporate both rasterized and vectorized HD map estimation within end-to-end training. UniAD~\cite{hu2023uniad} and OccNet~\cite{tong2023occnet}, for instance, approach online mapping as a dense prediction task, predicting the per-pixel or per-voxel semantics of map elements, whereas VAD produces vectorized HD map representations.
In each architecture, mapping serves as both an auxiliary training task and an internal static world representation that informs downstream components. While these methods yield the most integrated stacks, they only implicitly account for uncertainty. Accordingly, our work can be incorporated within end-to-end stacks to provide an explicit model of uncertainty and improve overall system performance.

\begin{figure*}[t]
  \centering
  \includegraphics[width=\linewidth]{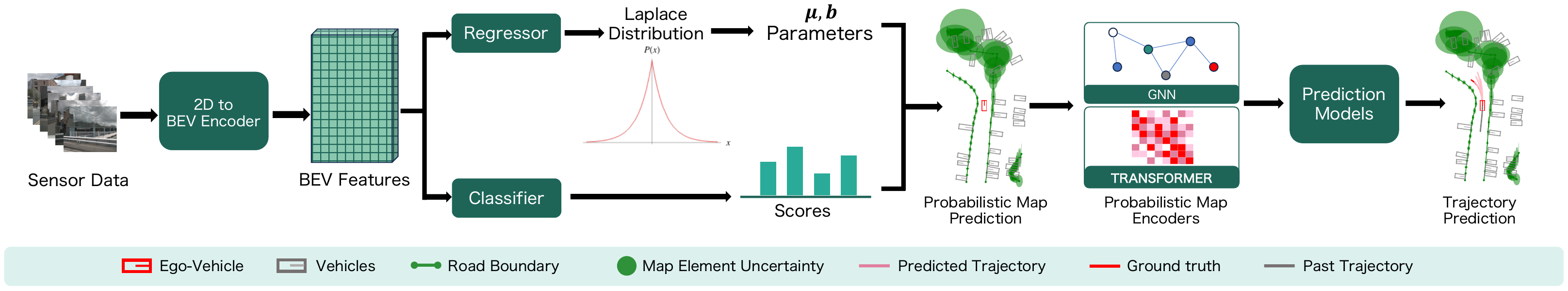}

\vspace{-0.2cm}
  
  \caption{Many online HD vector map estimation methods operate by encoding multi-camera images, transforming them to a common BEV feature space, and regressing map element vertices. Our work augments this common output structure with a probabilistic regression head, modeling each map vertex as a Laplace distribution. To assess the resulting downstream effects, we further extend downstream prediction models to encode map uncertainty, augmenting both GNN-based and Transformer-based map encoders.
  }
  \label{fig:diagram}

  \vspace{-0.5cm}
  
\end{figure*}

\section{Producing Online Map Uncertainty}
\label{sec:method_uncertainty}

As mentioned in \cref{sec:related_work}, there are many potential architectures for vectorized HD map estimation and accordingly many potential sources of uncertainty (e.g., perspective-to-BEV transformations, multi-sensor fusion, polyline vertex locations, element connectivity). However, as depicted in \cref{fig:diagram}, nearly all architectures utilize a point regression and classification head to predict the locations of polyline vertices and identify what kind of map element they are (e.g., lane line, stop line, road boundary), respectively. Thus, to ensure the general applicability of our proposed uncertainty formulation to a variety of map estimation approaches, we focus on extending this common output structure to additionally produce location and class uncertainty.

\textbf{Regression Uncertainty.} Vectorized HD map estimation models typically employ a simple MLP architecture for their regression heads. For each map element, the regression head produces a 2D vector representing normalized BEV $(x, y)$ coordinates. To transform this into a probabilistic model, we replace the regression head with one that additionally outputs uncertainty parameters associated with the predicted points. While a common choice is to assume Gaussian uncertainty, we found that this yields instabilities during training. Instead, we model each map element vertex $\mathbf{v} = (v_1, v_2)$ with two univariate Laplace distributions.

Accordingly, the 
joint probability density for a map element $M$ with $V$ vertices, denoted $M = \{\mathbf{v}^{(i)}\}_{i=1}^V$, is
\begin{equation}
    f(M \mid \boldsymbol{\mu}, \mathbf{b}) = \prod_{i=1}^{V} \prod_{j=1}^{2} \frac{1}{2b^{(i)}_j} \exp\left(-\frac{|v^{(i)}_j - \mu^{(i)}_j|}{b^{(i)}_j}\right),
\end{equation}
where $\mu^{(i)}_j \in \mathbb{R}$ and $b^{(i)}_j \in \mathbb{R}$ are the location and scale parameters of the Laplace distribution for the $j^\text{th}$ dimension of the $i^\text{th}$ vertex of the map element.

With its sharper peak and heavier tails, the Laplace distribution is particularly adept at handling outliers compared to the Gaussian distribution. Further, many online map estimation methods are trained using the Manhattan ($\ell_1$) distance as their regression loss~\cite{MapTR,maptrv2,yuan2024streammapnet}, making the Laplace distribution a natural choice. As we will show in \cref{sec:expt_prod_uncertainty}, such a direct uncertainty formulation can already model interesting sources of uncertainty, such as occlusions.

\textbf{Classification Uncertainty.} The classification head predicts class confidence scores for each regressed vertex. Since this head is already probabilistic (it is a Categorical distribution), we simply expose the semantic class logits to downstream consumers.

\textbf{Training Loss.} To train an uncertainty-producing map estimation model, only the regression loss $L_\text{R}$ needs to be changed to a Negative Log-Likelihood (NLL) loss,
\begin{equation}
    L_{\text{R}}(M \mid \boldsymbol{\mu}, \mathbf{b}) = \sum_{i=1}^{V} \sum_{j=1}^{2} \log\left(2b^{(i)}_j\right) + \frac{|v^{(i)}_j - \mu^{(i)}_j|}{b^{(i)}_j}.
\end{equation}

\textbf{Models.} In this work, we extend the MapTR~\cite{MapTR}, MapTRv2~\cite{maptrv2}, and StreamMapNet~\cite{yuan2024streammapnet} online HD map estimation models to demonstrate the benefits of producing map uncertainty. We choose these approaches as they are all very recent works that achieve state-of-the-art online HD mapping performance. At a high level, MapTR~\cite{MapTR} and MapTRv2~\cite{maptrv2} are Transformer-based models which adopt an encoder-decoder architecture. As depicted in \cref{fig:diagram}, they first encode RGB images to a common BEV feature $\mathcal{B} \in \mathbb{R}^{H \times W \times C}$ (using the LSS~\cite{philion2020lss}-based BEVPoolv2~\cite{huang2022bevpoolv2}). Their map decoders consist of map queries and several decoder layers. Each decoder layer utilizes self-attention and cross-attention to update the map queries before finally decoding them with a (non-probabilistic) regression and classification head. Note that, while MapTRv2~\cite{maptrv2} optionally supports LiDAR, we do not use it.

On the other hand, StreamMapNet~\cite{yuan2024streammapnet} focuses on operating from streaming data, containing an additional memory buffer that stores prior queries and BEV features which are ego-pose-corrected and combined with queries and BEV features in the current timestep to incorporate temporal information.

Each model produces three types of map elements: road boundary, pedestrian crosswalk, and lane divider. MapTRv2~\cite{maptrv2} can additionally output lane centerlines, which have been shown to be critical for trajectory forecasting~\cite{dauner2023parting}.
Each of these four models predict vectorized map elements within a perception range of $60m$ longitudinally by $30m$ laterally (centered on the AV).

\section{Incorporating Map Uncertainty in Trajectory Prediction}
\label{sec:method_incorporating}

The vast majority of trajectory prediction models employ an encoder-decoder architecture~\cite{RudenkoPalmieriEtAl2019}, where the encoder encodes scene context (e.g., vectorized map information and agent trajectories) and the decoder leverages such information to predict the future motion of surrounding agents. In the encoder, as depicted in \cref{fig:diagram}, map element vertices are most commonly encoded as nodes in a GNN (e.g., in DenseTNT~\cite{GuSunEtAl2021}) or tokens in Transformer cross-attention (e.g., in HiVT~\cite{zhou2022hivt}). In either of these models, vertex coordinates are first encoded by an MLP $\phi_{\text{v}}$ before being incorporated in message passing or attention layers. Formally, the $i^\text{th}$ vertex of a map element $M$ is encoded as $\mathbf{e}_{\text{v}}^{(i)} = \phi_{\text{v}}\left(\mathbf{v}^{(i)}\right)$.

To incorporate upstream uncertainty information in prediction models, we instead encode the Laplace distribution location $\boldsymbol{\mu}$ and scale $\mathbf{b}$ parameters, as well as the class probabilities $\mathbf{c} \in \Delta^{C-1}$, yielding
\begin{equation}\label{eq:incorporate}
    \mathbf{e}_{\text{v,unc}} = \phi_{\text{v}}^{(i)}\left([\boldsymbol{\mu}^{(i)}; \mathbf{b}^{(i)}; \mathbf{c}^{(i)}]\right),
\end{equation}
where $[\cdot;\cdot]$ represents concatenation and $\Delta^{C-1}$ denotes the probability simplex with $C$ classes.

\textbf{Models.} We augment the DenseTNT~\cite{GuSunEtAl2021} and HiVT~\cite{zhou2022hivt} trajectory prediction models to incorporate upstream map uncertainty, choosing these models as they implement the two dominant paradigms of encoding map information: GNNs and Transformers, respectively. 

At a high-level, DenseTNT~\cite{GuSunEtAl2021} leverages VectorNet~\cite{gao2020vectornet} to extract features from lanes and agents. It employs a hierarchical GNN consisting of two stages: local information from individual polylines is first aggregated and encoded, followed by global interactions between the resulting polyline node features. DenseTNT~\cite{GuSunEtAl2021} then employs a dense goal probability estimation technique to predict the endpoints of trajectories and generates complete trajectories based on the best goal candidates. To augment DenseTNT to incorporate map uncertainty, we integrate map element vertex uncertainty into the lane feature encoding, alongside the vertex coordinates (as in \cref{eq:incorporate}). These uncertainty-enhanced vectors are then encoded with VectorNet~\cite{gao2020vectornet}.

HiVT~\cite{GuSunEtAl2021} similarly encodes scene context in two hierarchical stages: first encoding local context (relative to each agent), followed by global interaction modeling between the local neighborhoods to capture long-range dependencies and scene-level dynamics. The resulting agent embeddings are then decoded with an MLP to produce the parameters of a multimodal trajectory distribution.

We augment HiVT~\cite{GuSunEtAl2021} to incorporate map uncertianty by inputting the estimated map as a \emph{point set}, instead of a vector set as in the original model, enabling the direct incorporation of vertex uncertainty in the encoder. Specifically, the uncertainty (scale parameter $b$) of each point is directly concatenated with the mean values of the point set, which are then encoded by the local neighborhood encoder together with agent trajectory information. 

As we will show in \cref{sec:expt_incorporating}, incorporating polyline uncertainty directly in this manner enables prediction models to understand when map element estimations may be unreliable and adjust their outputs accordingly, yielding significant accuracy improvements.
\section{Experiments}
\label{sec:expt}

\subsection{Experiment Setup}
\label{sec:setup}
\textbf{Dataset.} 
We evaluate our probabilistic map estimation and prediction framework on the large-scale nuScenes dataset~\cite{CaesarBankitiEtAl2019}, which provides ground truth (GT) HD maps, sensor data (RGB images), as well as agent trajectories. It consists of 1000 driving scenes with each scene sampled at 2Hz, and is split into training, validation, and test sets containing 500, 200, and 150 scenes, respectively.

We leverage trajdata~\cite{ivanovic2023trajdata} to provide a unified interface between vectorized map estimation models and downstream prediction models. To ensure compatibility across prediction models, we upsample nuScenes' data frequency to 10Hz (from its original 2Hz) using trajdata's time interpolation utilities~\cite{ivanovic2023trajdata}. This modification provides a denser dataset, thereby facilitating finer-grained analyses and aligning our data more closely with the real-time execution rates of onboard prediction models. Finally, we task each prediction model to predict motion 3 seconds into the future from 2 seconds of history. 

\textbf{Metrics.} The Chamfer distance $D_{\text{Ch}}$ is employed to measure the distance between two maps (represented as point sets $S_1$ and $S_2$). Formally,
\begin{equation}
    D_{\text{Ch}} = \sum_{x \in S_1} \min_{y \in S_2} \frac{\| x - y \|_2}{|S_1|} + \sum_{y \in S_2} \min_{x \in S_1} \frac{\| y - x \|_2}{|S_2|}.
\end{equation}
In line with prior works~\cite{MapTR,maptrv2,yuan2024streammapnet}, we adopt Average Precision (AP) as the evaluation metric for our probabilistic map construction of four map elements: road boundary, pedestrian crossing, lane divider, and lane centerlines. Mean AP (mAP) is further calculated as the mean AP under three distinct $D_{\text{Ch}}$ thresholds: 0.5 m, 1.0 m, and 1.5 m.

\begin{figure*}[t]
\centering

\begin{minipage}[b]{.24\textwidth}
    \includegraphics[width=\linewidth]{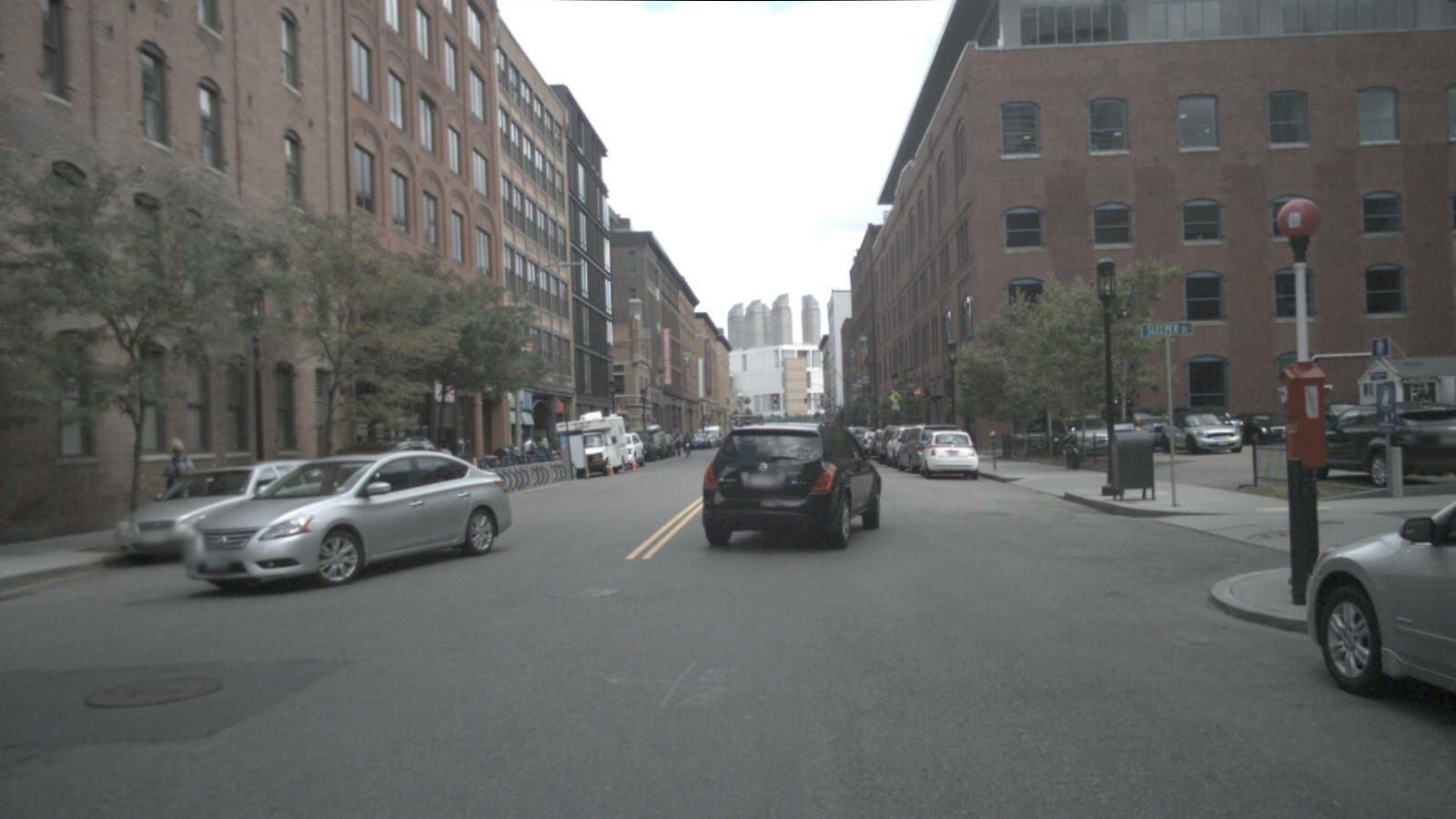}
    \includegraphics[width=\linewidth]{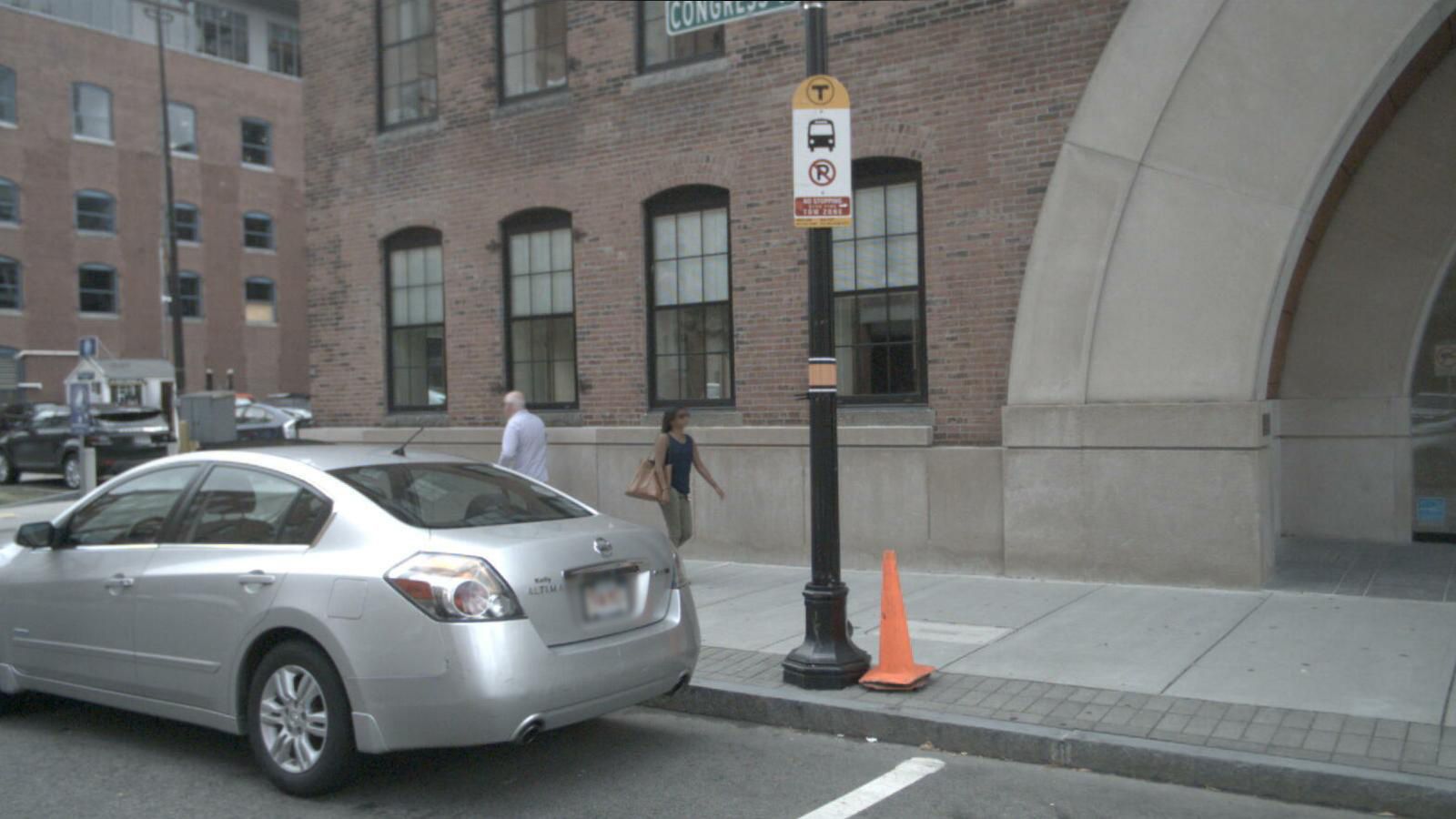}
\end{minipage}%
\hfill 
\begin{minipage}[b]{.74\textwidth}
  \centering
  \begin{subfigure}{.19\linewidth}
    \includegraphics[width=\linewidth]{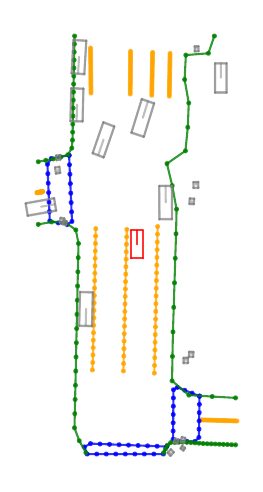}
    \caption{GT}
  \end{subfigure}%
  \begin{subfigure}{.19\linewidth}
    \includegraphics[width=\linewidth]{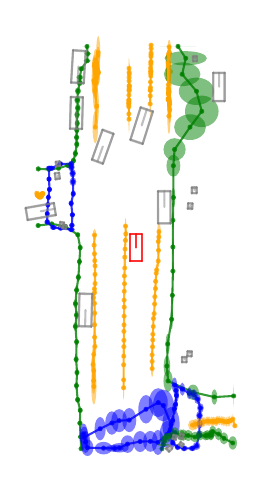}
    \caption{MapTR}
  \end{subfigure}%
  \begin{subfigure}{.19\linewidth}
    \includegraphics[width=\linewidth]{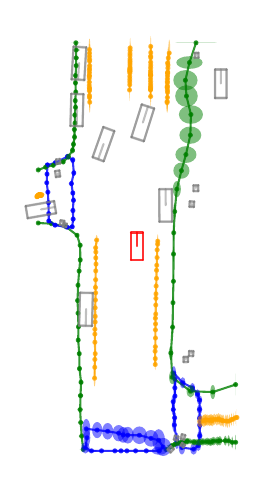}
    \caption{MapTRv2}
  \end{subfigure}
  \begin{subfigure}{.19\linewidth}
    \includegraphics[width=\linewidth]{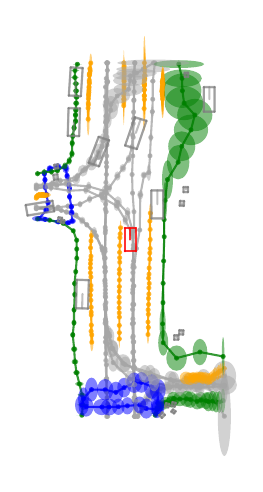}
    \caption{MapTRv2-Center}
  \end{subfigure}
  \begin{subfigure}{.19\linewidth}
    \includegraphics[width=\linewidth]{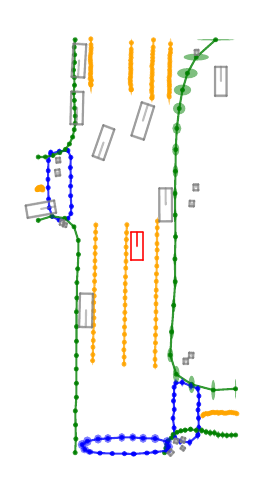}
    \caption{StreamMapNet}
    \label{fig:occlusion1_stream}
  \end{subfigure}%
\end{minipage}

\vspace{-0.2cm}

\caption{Our proposed uncertainty formulation is able to capture uncertainty stemming from occlusions between the AV's cameras and surrounding map elements. \textbf{Left:} Images from the front and front-right cameras. \textbf{Right:} HD maps from our augmented online HD mapping models. Ellipses show the std.\ dev.\ of distributions. Colors are {\color{darkgreen} road boundary}, {\color{darkyellow} lane divider}, {\color{blue} pedestrian crossing}, {\color{gray} lane centerline}.}
\label{fig:occlusion1}

\vspace{-0.4cm}

\end{figure*}

\begin{figure*}[htbp]
\centering

\begin{minipage}[b]{.24\textwidth}
    \includegraphics[width=\linewidth]{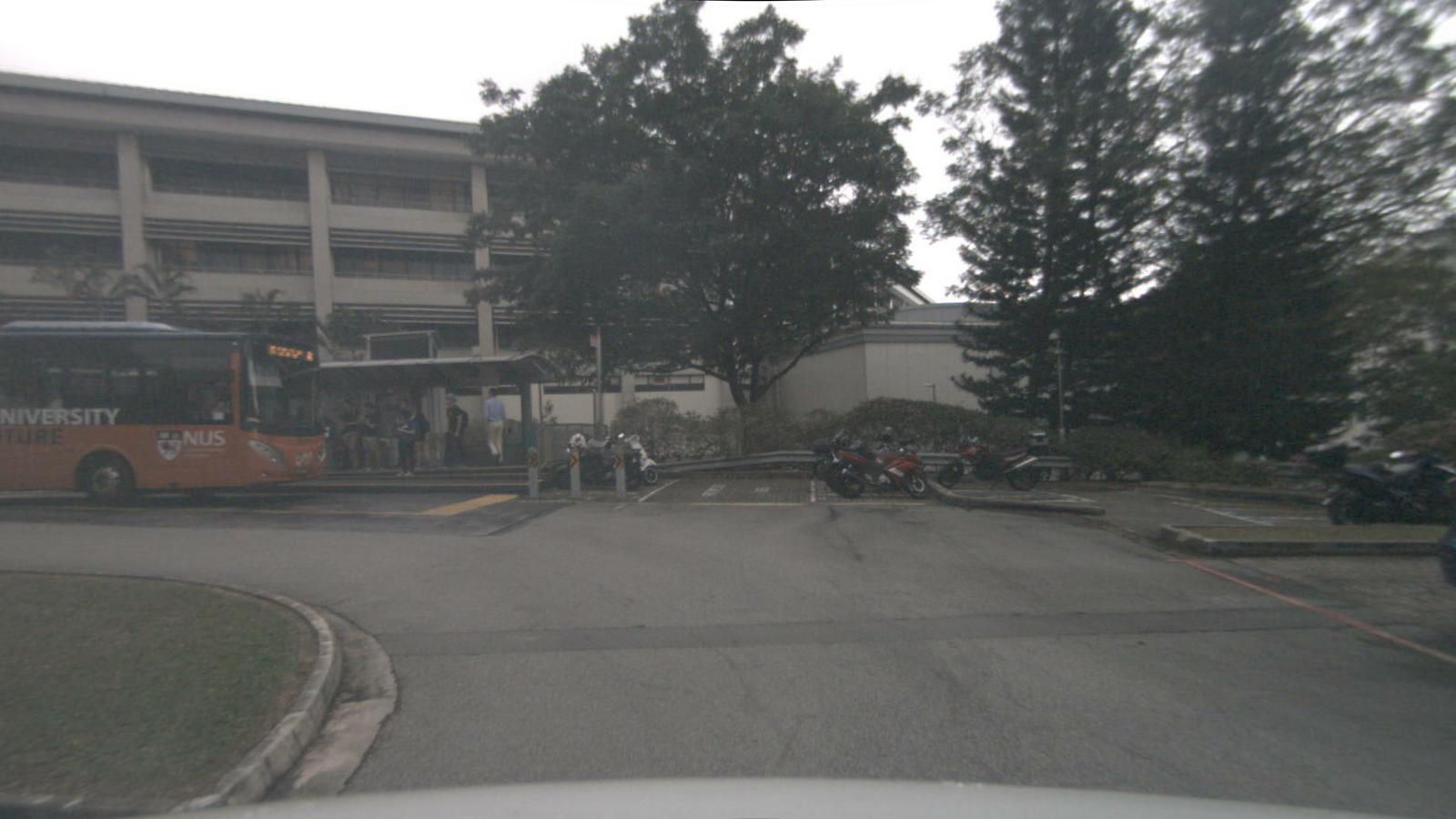}
    \includegraphics[width=\linewidth]{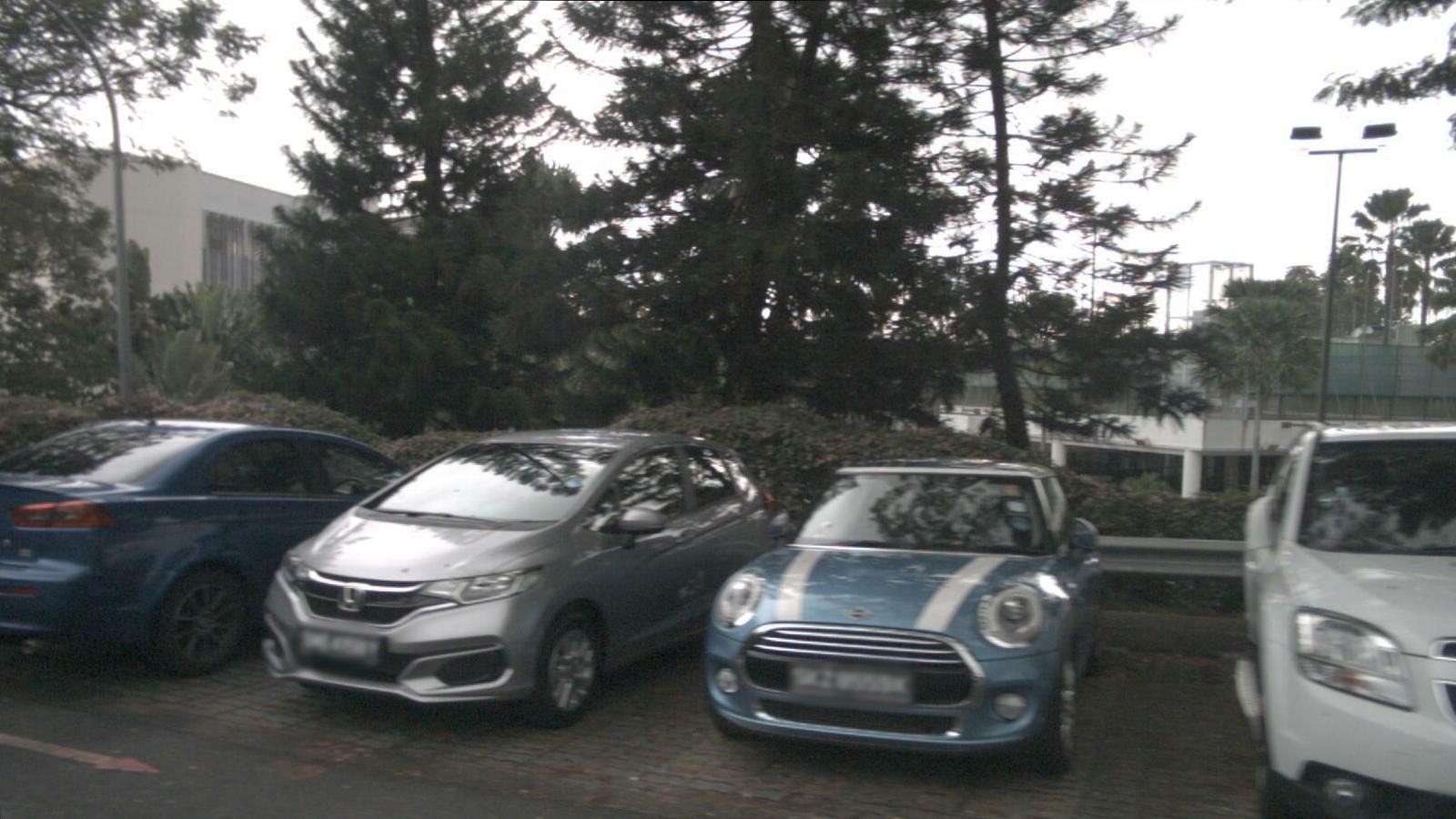}
\end{minipage}%
\hfill 
\begin{minipage}[b]{.74\textwidth}
  \centering
  \begin{subfigure}{.19\linewidth}
    \includegraphics[width=\linewidth]{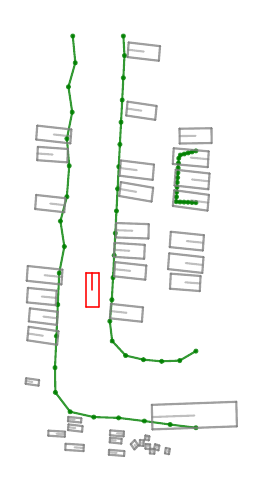}
    \caption{GT}
  \end{subfigure}%
  \begin{subfigure}{.19\linewidth}
    \includegraphics[width=\linewidth]{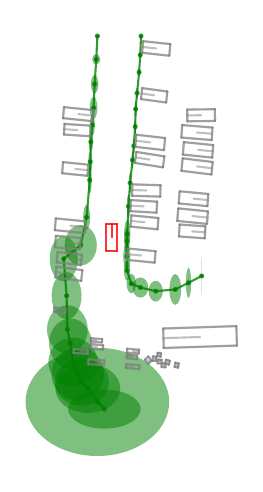}
    \caption{MapTR}
  \end{subfigure}%
  \begin{subfigure}{.19\linewidth}
    \includegraphics[width=\linewidth]{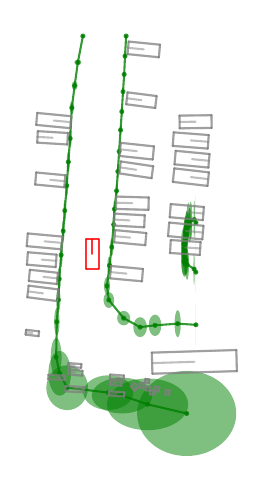}
    \caption{MapTRv2}
  \end{subfigure}
  \begin{subfigure}{.19\linewidth}
    \includegraphics[width=\linewidth]{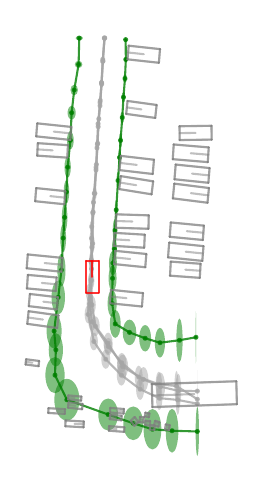}
    \caption{MapTRv2-Center}
  \end{subfigure}
  \begin{subfigure}{.19\linewidth}
    \includegraphics[width=\linewidth]{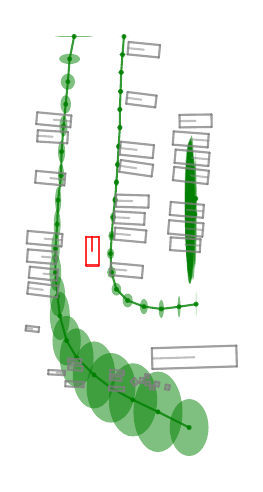}
    \caption{StreamMapNet}
  \end{subfigure}%
\end{minipage}
\caption{In a dense parking lot, many models fail to produce accurate maps. \textbf{Left:} Images from the rear and rear-left cameras. \textbf{Right:} HD maps from our augmented online HD mapping models. Ellipses show the std.\ dev.\ of distributions. Colors indicate {\color{darkgreen} road boundary}, {\color{darkyellow} lane divider}, {\color{blue} pedestrian crossing}, {\color{gray} lane centerline}.}
\label{fig:occlusion2}

\vspace{-0.5cm}

\end{figure*}

For trajectory prediction, we evaluate our model on standard metrics adopted by numerous recent prediction challenges~\cite{ChangLambertEtAl2019,wilson2021argoverse2,waymo_open_motion_dataset}, specifically minimum Average Displacement Error (minADE), minimum Final Displacement Error (minFDE), and Miss Rate (MR)~\cite{ChangLambertEtAl2019}. For each agent, 6 potential trajectories are output for evaluation. The minADE metric computes the average Euclidean ($\ell_2$) distance in meters across all future time steps between the most accurately predicted trajectory and the ground truth trajectory. Similarly, minFDE calculates the error of only the final predicted time step. The most accurately predicted trajectory is identified based on having the smallest FDE. MR quantifies the proportion of scenarios where the endpoint of the best-predicted trajectory deviates from the ground truth trajectory's endpoint by more than $2.0$ meters.

\textbf{Data Preprocessing and Training.} We standardize all agent and lane features by transforming their coordinates to be relative to ego-vehicle's position, as well as rotating the scene to make the AV's heading point up. As a consequence, we also transform the map uncertainty with
\begin{equation}
\begin{split}
\sigma_{x'} = \sqrt{\sigma_x^2\cos^2(\theta)+\sigma_y^2\sin^2(\theta)}, \\
\sigma_{y'} = \sqrt{\sigma_x^2\sin^2(\theta)+\sigma_y^2\cos^2(\theta)}, 
\end{split}
\end{equation}
where $\theta$ is the rotated angle and $\sigma = \sqrt{2} \cdot b$ is the Laplace distribution's standard deviation (derived from its scale parameter $b$). All models are trained using a single NVIDIA GeForce RTX 4090 GPU.
For full model hyperparameter settings and training details, please refer to \cref{sec:supp_training}. 

\subsection{Producing Map Uncertainty}
\label{sec:expt_prod_uncertainty}



Augmenting MapTR~\cite{MapTR}, MapTRv2~\cite{maptrv2} (and its centerline-producing version), and StreamMapNet~\cite{yuan2024streammapnet} to produce uncertainty does not substantially affect their original mapping performance. We are able to reproduce most models' published performance within 2\% mAP, with some uncertainty-augmented versions even outperforming the original models. In the following, we analyze the uncertainty output by these map estimation models and identify various sources of uncertainty that our approach captures.


\textbf{Uncertainty from Occlusion.} Our proposed uncertainty formulation is able to capture uncertainty stemming from occlusions between the AV's cameras and the surrounding map elements. As can be seen in \cref{fig:occlusion1}, the top right portion of the map (forward and to the right of the AV) is occluded by a red callbox and a grey parked car.
Importantly, even though our work only modifies the final output heads, \cref{fig:occlusion1} shows that it is still able to identify when certain map elements are occluded in the input RGB images.

We also observe in \cref{fig:occlusion1_stream} the benefits of StreamMapNet's memory module: It outputs less uncertainty in the same top-right portion of the map, owing to its incorporation of temporal information from past frames (when map elements were visible). Conversely, MapTR and MapTRv2 are single-frame models and cannot enjoy such benefits.


Similarly, \cref{fig:occlusion2} visualizes a scenario where all models struggle to delineate between driveable road and parking spots in a parking lot (region at the bottom of the map, behind the ego-vehicle). Additionally, in easily-observed parts of the map (the region at the top of the map, in front of the ego-vehicle), all models produce confident predictions with very little uncertainties.


\begin{figure*}[htbp]
\centering
\includegraphics[width=\textwidth]{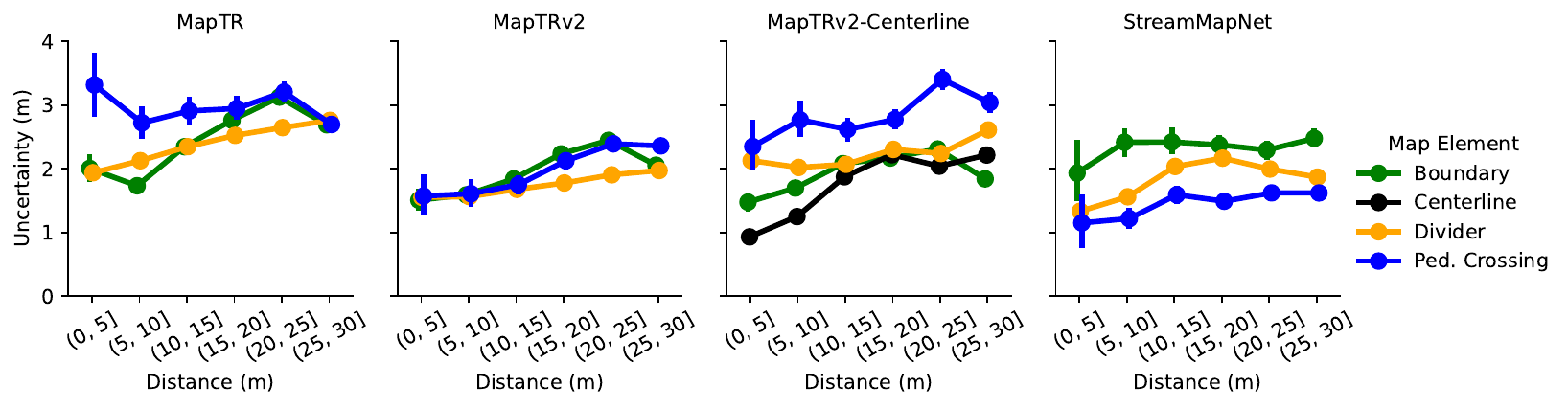}

\vspace{-0.5cm}

\caption{Our uncertainty formulation captures the fact that uncertainty generally increases with the distance between the predicted map elements and the AV, owing to the difficulty of resolving the details of faraway objects in images. Error bars show 95\% confidence intervals.}
\label{fig:distance}

\vspace{-0.3cm}

\end{figure*}

\begin{figure*}[htbp]
\centering
\includegraphics[width=\textwidth]{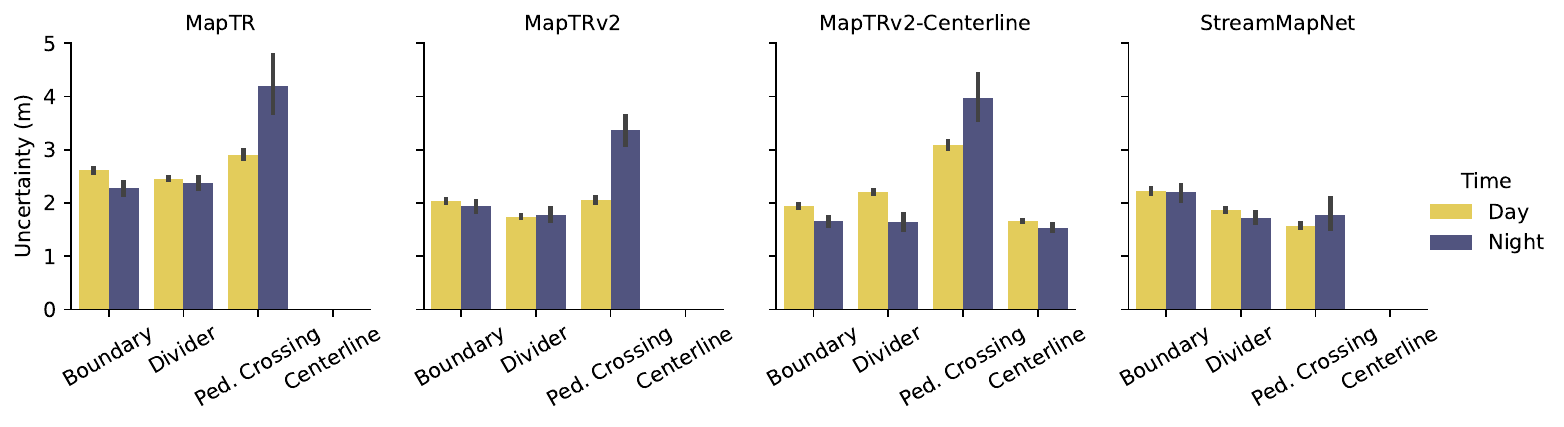}

\vspace{-0.5cm}

\caption{Different time-of-day lighting can significantly affect the confidence with which map estimation models predict certain elements, such as pedestrian crossings. Error bars show 95\% confidence intervals.}
\label{fig:night}

\vspace{-0.5cm}

\end{figure*}

\textbf{Uncertainty from Sensor Range.} Another important source of uncertainty in map estimation is the distance from the onboard cameras to the map elements, stemming from the 2D-to-BEV transformation in many mapping models. As can be seen in \cref{fig:distance}, map uncertainty generally increases with increasing distance between the vehicle and the corresponding map elements. We can also see that MapTRv2 generally yields lower uncertainties than its predecessor MapTR, matching the fact that MapTRv2 is generally more accurate than MapTR~\cite{maptrv2}. StreamMapNet's uncertainty remains relatively constant compared to the other per-frame models, again highlighting the benefits of aggregating temporal information.

Further, note the increase in pedestrian crosswalk uncertainty when MapTRv2 is tasked with estimating lane centerlines. One hypothesis is that lane centerlines frequently pass through pedestrian crosswalks, causing confusion in the model about which polyline to optimize during training.

\textbf{Uncertainty from Lighting and Weather.} \cref{fig:night} and \cref{fig:rain} in \cref{sec:supp_vis} show the effect of different lighting and weather conditions, respectively, on map estimation uncertainty. 
In \cref{fig:night}, we can see that map elements which are typically lit by street lamps, vehicle headlights, and/or contain reflective surfaces (i.e., lane boundaries and dividers) retain the same level of uncertainty in day and nighttime. Conversely, models predict pedestrian crossings with significantly more uncertainty at night, potentially indicating that they may not have the same consistent lighting at night compared to other elements.
\cref{fig:rain} in \cref{sec:supp_vis} additionally shows that StreamMapNet produces more uncertainty in rainy conditions, indicating potential difficulties in aggregating temporal information due to rain.

\textbf{Uncertainty from Motion.} Finally, \cref{fig:speed} in \cref{sec:supp_vis} shows that current models do not have any particular lack of confidence across different AV driving speeds. However, nuScenes~\cite{CaesarBankitiEtAl2019} does not contain much high-speed driving (shown in Figure 9 of~\cite{ivanovic2023trajdata}), leaving high-speed analyses (e.g., about rolling shutter effects) to future work.

\subsection{Incorporating Map Uncertainty in Prediction}
\label{sec:expt_incorporating}

To evaluate the effect of incorporating map uncertainty in downstream autonomy stack components, we train DenseTNT~\cite{GuSunEtAl2021} and HiVT~\cite{zhou2022hivt} on the outputs of the aforementioned mapping models with and without our output uncertainty formulation, yielding 16 total combinations.

\begin{table*}
  \centering
  \resizebox{1\linewidth}{!}{
  \begin{tabular}{@{}l|lll|lll@{}}
    \toprule
    Prediction Method & \multicolumn{3}{c|}{HiVT~\cite{zhou2022hivt}} & \multicolumn{3}{c}{DenseTNT~\cite{GuSunEtAl2021}} \\
    \midrule
    Online HD Map Method & minADE $\downarrow$ & minFDE $\downarrow$ & MR $\downarrow$ & minADE $\downarrow$ & minFDE $\downarrow$ & MR $\downarrow$ \\
    \midrule
    MapTR~\cite{MapTR} & 0.4015 & 0.8418 & 0.0981 & 1.091&2.058&0.3543 \\
    MapTR~\cite{MapTR} + Ours & 0.3854 {\small \color{darkpastelgreen} ($-4\%$)} & 0.7909 {\small \color{darkpastelgreen} ($-6\%$)} & 0.0834 {\small \color{darkpastelgreen} ($\mathbf{-15}\%$)} & 1.089 {\small \color{gray} ($0\%$)} & 2.006 {\small \color{darkpastelgreen} ($-3\%$)} & 0.3499 {\small \color{darkpastelgreen} ($-1\%$)} \\
    \midrule
    MapTRv2~\cite{maptrv2} & 0.4057 & 0.8499 & 0.0992 &1.214 & 2.312 & 0.4138 \\ 
    MapTRv2~\cite{maptrv2} + Ours & 0.3930 {\small \color{darkpastelgreen} ($-3\%$)} & 0.8127 {\small \color{darkpastelgreen} ($-4\%$)} & 0.0857 {\small \color{darkpastelgreen} ($\mathbf{-14}\%$)} & 1.262 {\small \color{orange} ($+4\%$)} &2.340 {\small \color{orange} ($+1\%$)} &0.3912 {\small \color{darkpastelgreen} ($-5\%$)} \\
    \midrule
    MapTRv2-Centerline~\cite{maptrv2} & 0.3790 & 0.7822 & 0.0853 & 0.8466 & 1.345 & 0.1520  \\
    MapTRv2-Centerline~\cite{maptrv2} + Ours & 0.3727 {\small \color{darkpastelgreen} ($-2\%$)} & 0.7492 {\small \color{darkpastelgreen} ($-4\%$)} & 0.0726 {\small \color{darkpastelgreen} ($\mathbf{-15}\%$)} & 0.8135 {\small \color{darkpastelgreen} ($-4\%$)} & 1.311 {\small \color{darkpastelgreen} ($-3\%$)} & 0.1593 {\small \color{orange} ($+5\%$)} \\
    \midrule
    StreamMapNet~\cite{yuan2024streammapnet}  & 0.3972 & 0.8186 & 0.0926 & 0.9492 & 1.740 & 0.2569 \\
    StreamMapNet~\cite{yuan2024streammapnet}  + Ours & 0.3848 {\small \color{darkpastelgreen} ($-3\%$)} & 0.7954 {\small \color{darkpastelgreen} ($-3\%$)} & 0.0861 {\small \color{darkpastelgreen} ($-7\%$)} & 0.9036 {\small \color{darkpastelgreen} ($-5\%$)} & 1.645 {\small \color{darkpastelgreen} ($-5\%$)} & 0.2359 {\small \color{darkpastelgreen} ($-8\%$)} \\
    \bottomrule
  \end{tabular}
  }

\vspace{-0.3cm}
  
  \caption{Quantitative prediction results for all 16 mapping/prediction model combinations on the nuScenes~\cite{CaesarBankitiEtAl2019} dataset. In general, incorporating upstream map uncertainty improves the performance of prediction models, especially for endpoint prediction accuracy.}
  \label{tab:quant}

  \vspace{-0.5cm}
  
\end{table*}

\textbf{Prediction Accuracy Improvements.} As shown in \cref{tab:quant}, for virtualy all mapping/prediction model combinations, incorporating uncertainty yields better prediction performance. In general, the improvements in MR are the greatest, indicating that, by incorporating map uncertainty, 
prediction models can effectively adjust their behaviors to more closely match the ground truth future, especially at the endpoints. Endpoint accuracy is particularly important for trajectory prediction as many methods adopt a two-stage pipeline where the first stage predicts possible endpoints.

Further, although MapTRv2 significantly outperforms MapTR in map estimation~\cite{maptrv2}, there is little resulting difference in prediction performance (in fact, MapTR yields \emph{better} prediction performance than MapTRv2, see the first two sets of rows in \cref{tab:quant}). This indicates that accuracy improvements in upstream map estimation models may not directly improve downstream prediction accuracy.

The best prediction performance across all metrics (by far, in some cases) is achieved when leveraging the lane centerlines output by MapTRv2-Centerline. This confirms the superiority of using centerlines to guide trajectory prediction~\cite{dauner2023parting} and indicates where integrated systems can see the most improvement from future map estimation research.

Most interestingly, the performance of DenseTNT trained on maps from MapTRv2-Centerline exceeds the performance of DenseTNT trained on \emph{GT lane centerlines} (\cref{tab:better_than_gt}). The reason for this stems from MapTRv2-Centerline sometimes producing multiple centerlines for one lane. For a target-based model such as DenseTNT, multiple centerlines in the same lane provides a richer set of options for endpoint selection, focusing more closely the resulting endpoints within lanes and yielding better prediction performance.

\begin{table}
    \centering
    \begin{tabular}{l|cl}
     \toprule
     DenseTNT~\cite{GuSunEtAl2021} Training &\multicolumn{2}{|c}{Epochs to Convergence} \\ \midrule
     Map Model & Without Unc. & With Unc.  \\ \midrule
MapTR~\cite{MapTR}	&$8$& $4$ {\small \color{darkpastelgreen} ($\mathbf{-50}\%$)} \\
MapTRv2~\cite{maptrv2}		&$7$	&$4$ {\small \color{darkpastelgreen} ($\mathbf{-43}\%$)}\\
MapTRv2-Centerline~\cite{maptrv2}& $9$ &7	{\small \color{darkpastelgreen} ($\mathbf{-22}\%$)}\\
StreamMapNet~\cite{yuan2024streammapnet}		&$6$	&$4$	{\small \color{darkpastelgreen} ($\mathbf{-33}\%$)} \\
\bottomrule
    \end{tabular}

\vspace{-0.3cm}
    
    \caption{When trained with map uncertainty, DenseTNT~\cite{GuSunEtAl2021} consistently converges faster, arriving at equal or better validation performance, irrespective of the upstream mapping model.}
        \label{tab:training}

        \vspace{-0.2cm}
        
\end{table}

\textbf{Training Convergence Improvements.} Immediately during training, we found that all trajectory prediction models converge significantly faster when incorporating map uncertainty. As can be seen in \cref{tab:training}, DenseTNT trains to convergence much more quickly when incorporating map uncertainty, achieving optimal validation performance 2-4 epochs earlier than when only incorporating coordinates. 

\begin{table}
\setlength{\tabcolsep}{3pt}
\centering
\resizebox{1\linewidth}{!}{
    \begin{tabular}{l|lll}
         \toprule DenseTNT~\cite{GuSunEtAl2021} + Map Model &minADE $\downarrow$ & minFDE $\downarrow$ & MR $\downarrow$\\\midrule
GT Map & 0.8809& 1.489 & 0.1903\\
	MapTRv2-Centerline~\cite{maptrv2} &	0.8466 {\small \color{darkpastelgreen} ($-4\%$)} & 1.345 {\small \color{darkpastelgreen} ($\mathbf{-10}\%$)} &	0.1520 {\small \color{darkpastelgreen} ($\mathbf{-20}\%$)} \\
	MapTRv2-Centerline~\cite{maptrv2} + Ours&	0.8135 {\small \color{darkpastelgreen} ($-8\%$)} & 1.311 {\small \color{darkpastelgreen} ($\mathbf{-12}\%$)} & 0.1593 {\small \color{darkpastelgreen} ($\mathbf{-16}\%$)} \\
 \bottomrule
    \end{tabular}
    }

\vspace{-0.3cm}
    
    \caption{DenseTNT~\cite{GuSunEtAl2021} is able to achieve better prediction performance with MapTRv2-Centerline~\cite{maptrv2} compared to the GT map.}
    \label{tab:better_than_gt}

    \vspace{-0.5cm}
    
\end{table}

\begin{figure*}[htbp]
\centering

\begin{minipage}[b]{.19\textwidth}
    \includegraphics[width=\linewidth]{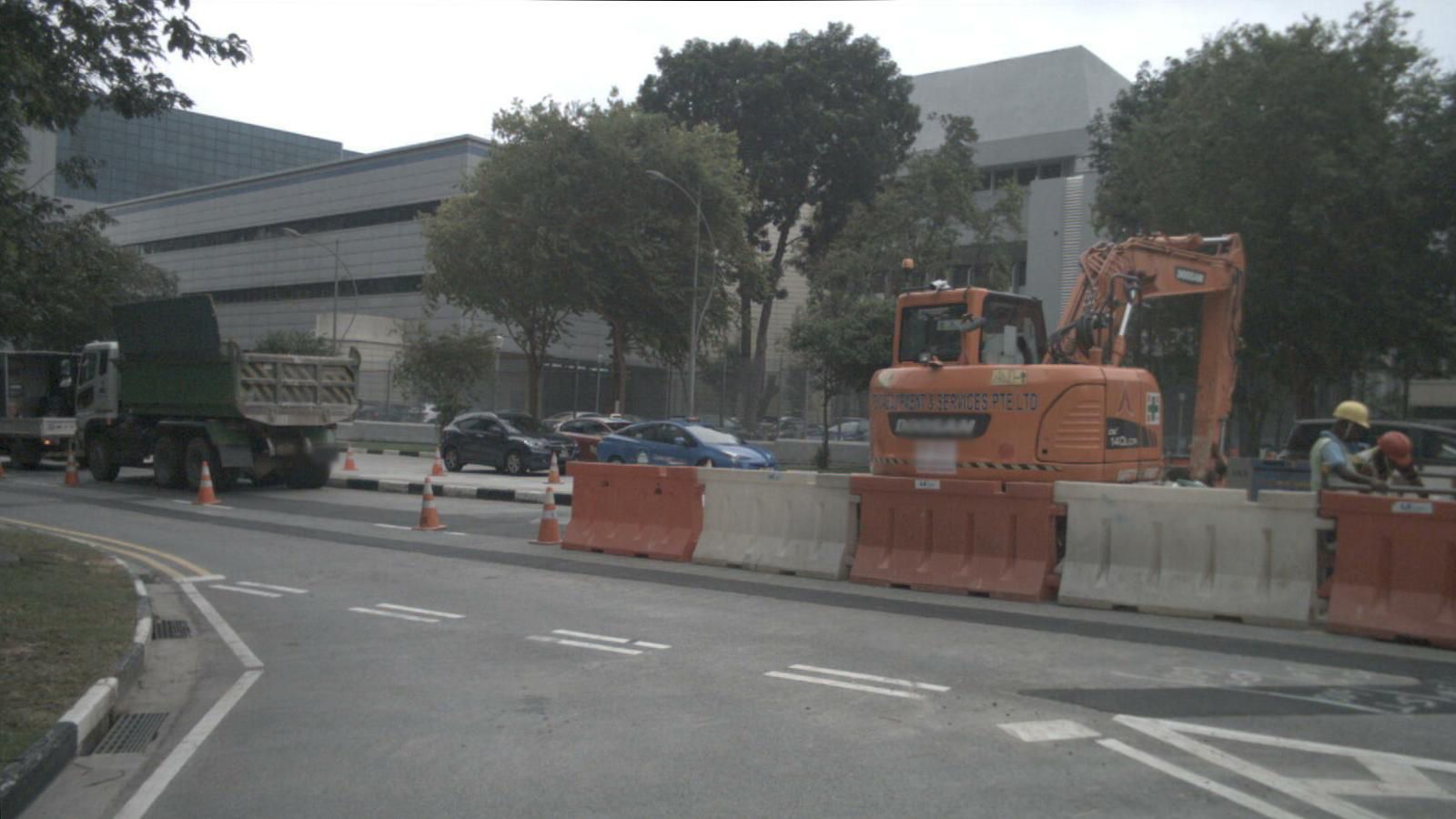}
    \includegraphics[width=\linewidth]{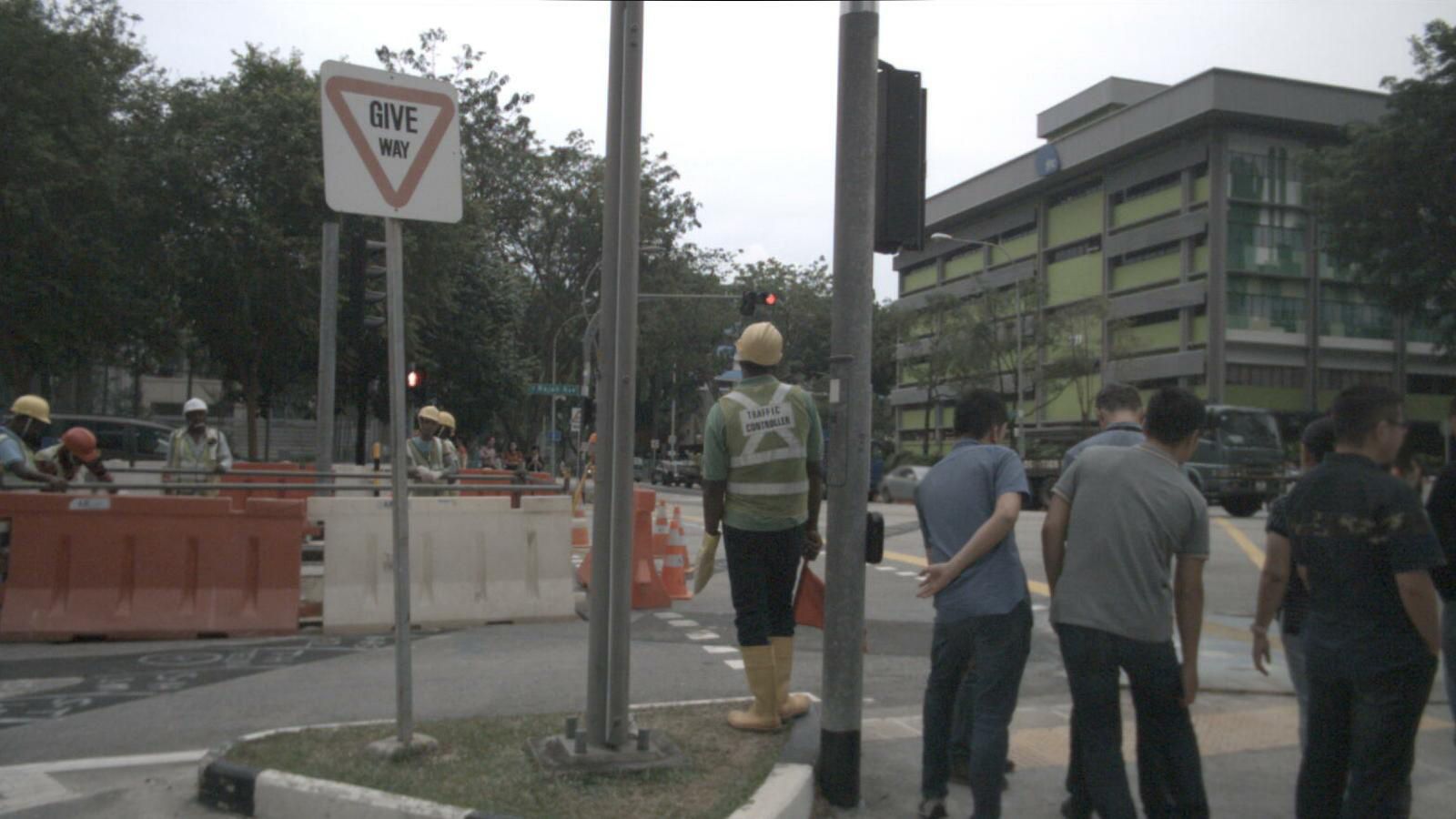}
    \includegraphics[width=\linewidth]{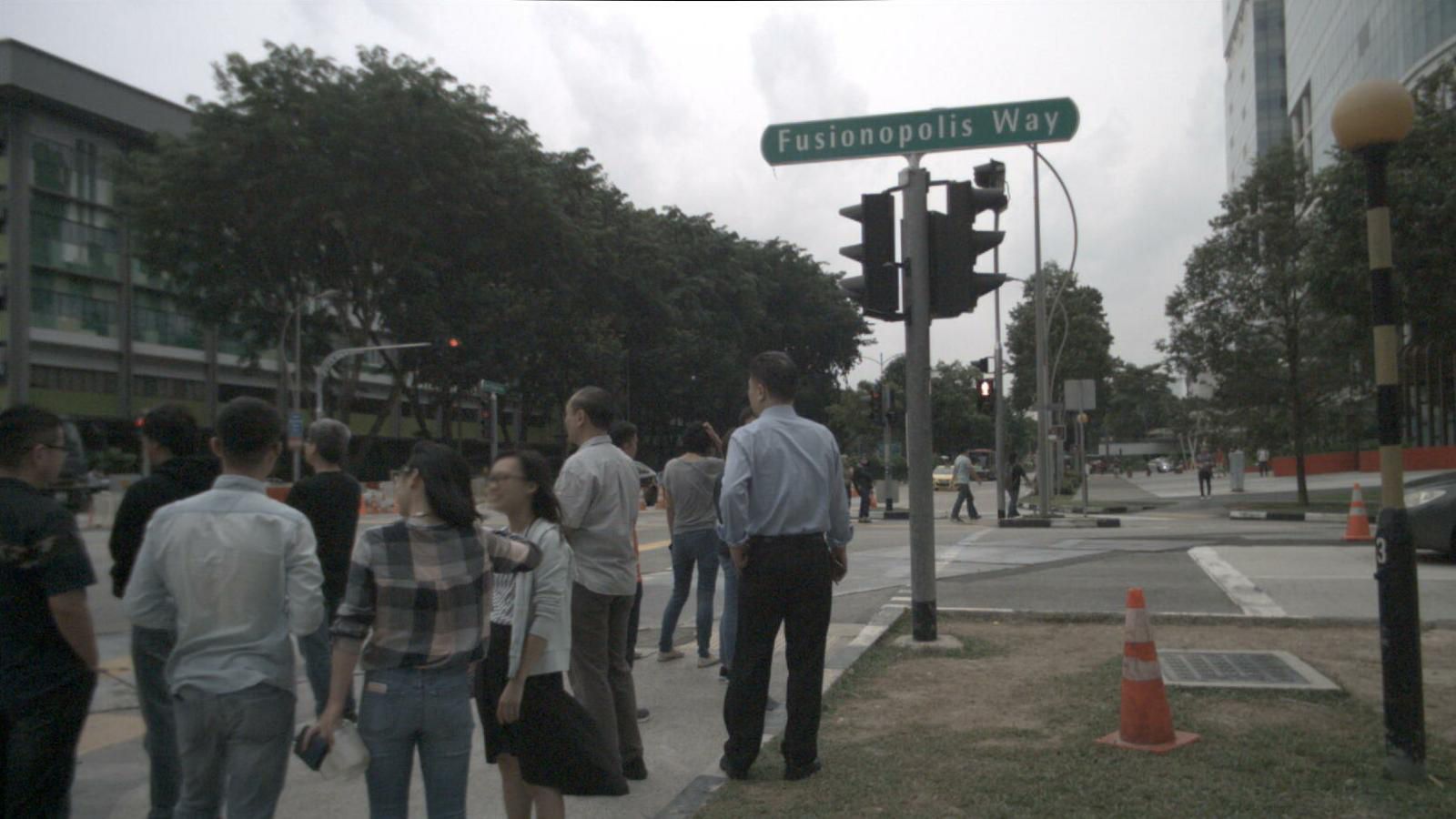}
\end{minipage}%
\hfill 
\begin{minipage}[b]{.79\textwidth}
  \centering
  \begin{subfigure}{.19\linewidth}
    \includegraphics[width=\linewidth]{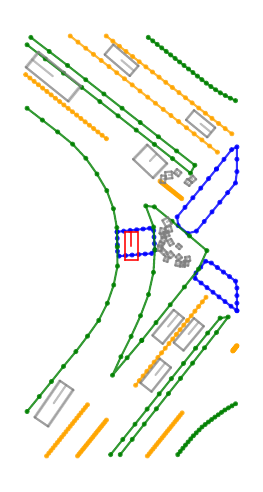}
    \caption{GT}
    \label{fig:vis1_gt}
  \end{subfigure}%
  \begin{subfigure}{.19\linewidth}
    \includegraphics[width=\linewidth]{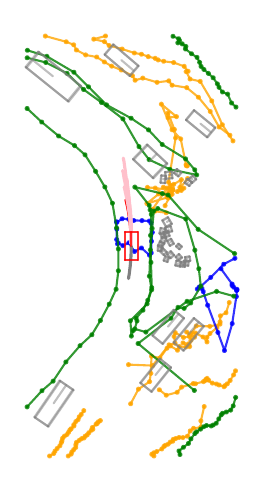}
    \caption{HiVT+MapTR}
    \label{fig:vis1_hmn}
  \end{subfigure}%
  \begin{subfigure}{.19\linewidth}
    \includegraphics[width=\linewidth]{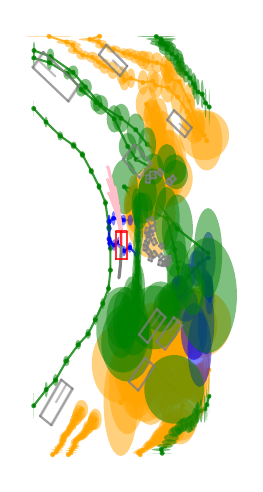}
    \caption{HiVT+MapTR}
    \label{fig:vis1_hmu}
  \end{subfigure}%
  \begin{subfigure}{.19\linewidth}
    \includegraphics[width=\linewidth]{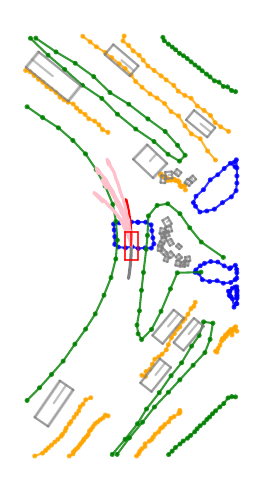}
    \caption{DenseTNT+Stream}
    \label{fig:vis1_dsn}
  \end{subfigure}
  \begin{subfigure}{.19\linewidth}
    \includegraphics[width=\linewidth]{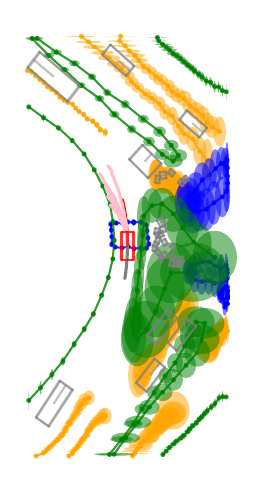}
    \caption{DenseTNT+Stream}
    \label{fig:vis1_dsu}
  \end{subfigure}
\end{minipage}
\caption{A complicated intersection with many map elements. By leveraging uncertainty information, both combinations of map estimation and prediction models show enhancements in prediction, correctly predicting that the center vehicle will stay in its current lane.}
\label{fig:vis1}


\end{figure*}

\begin{figure*}[htbp]
\centering

\begin{minipage}[b]{.19\textwidth}
    \includegraphics[width=\linewidth]{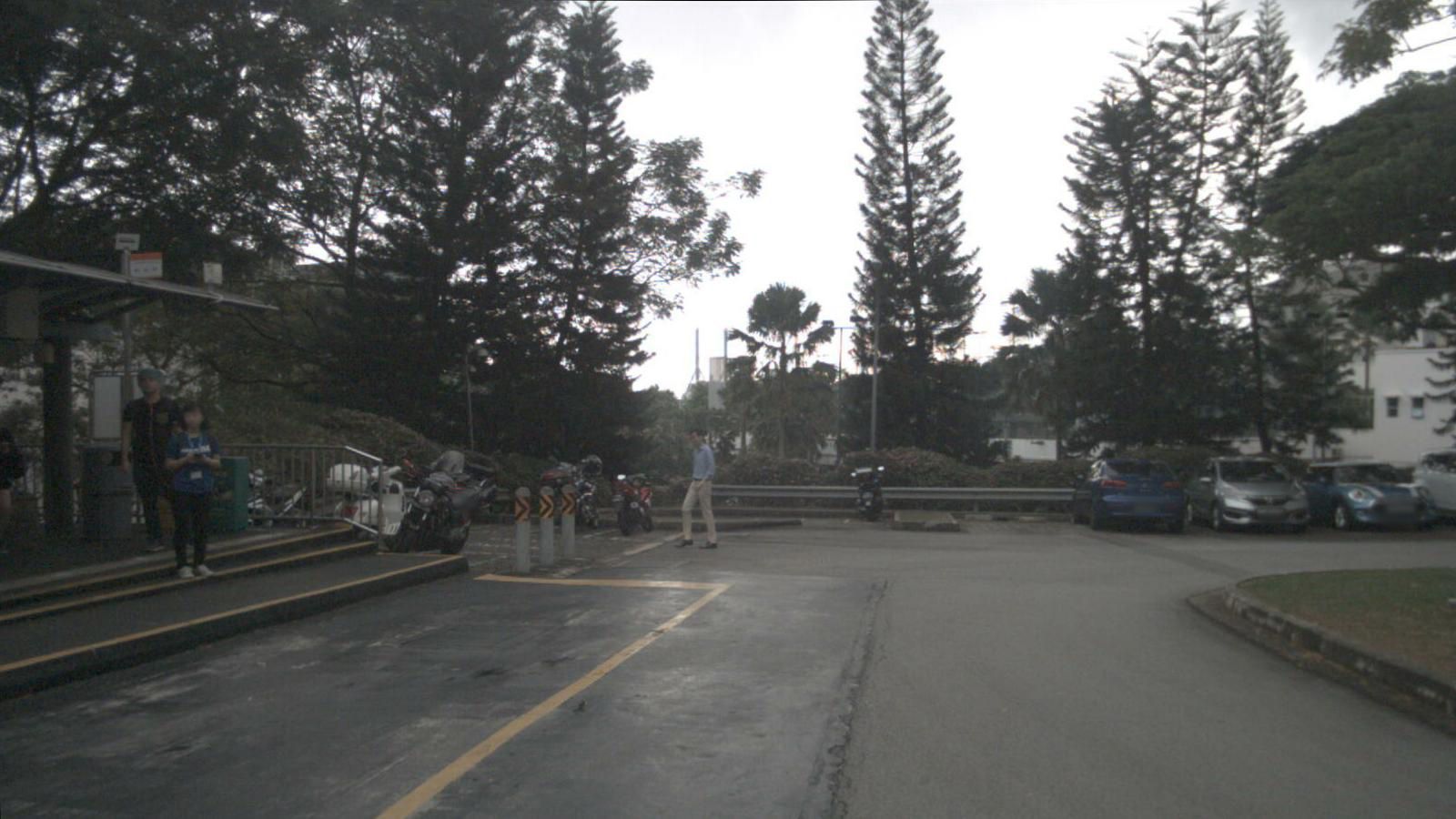}
    \includegraphics[width=\linewidth]{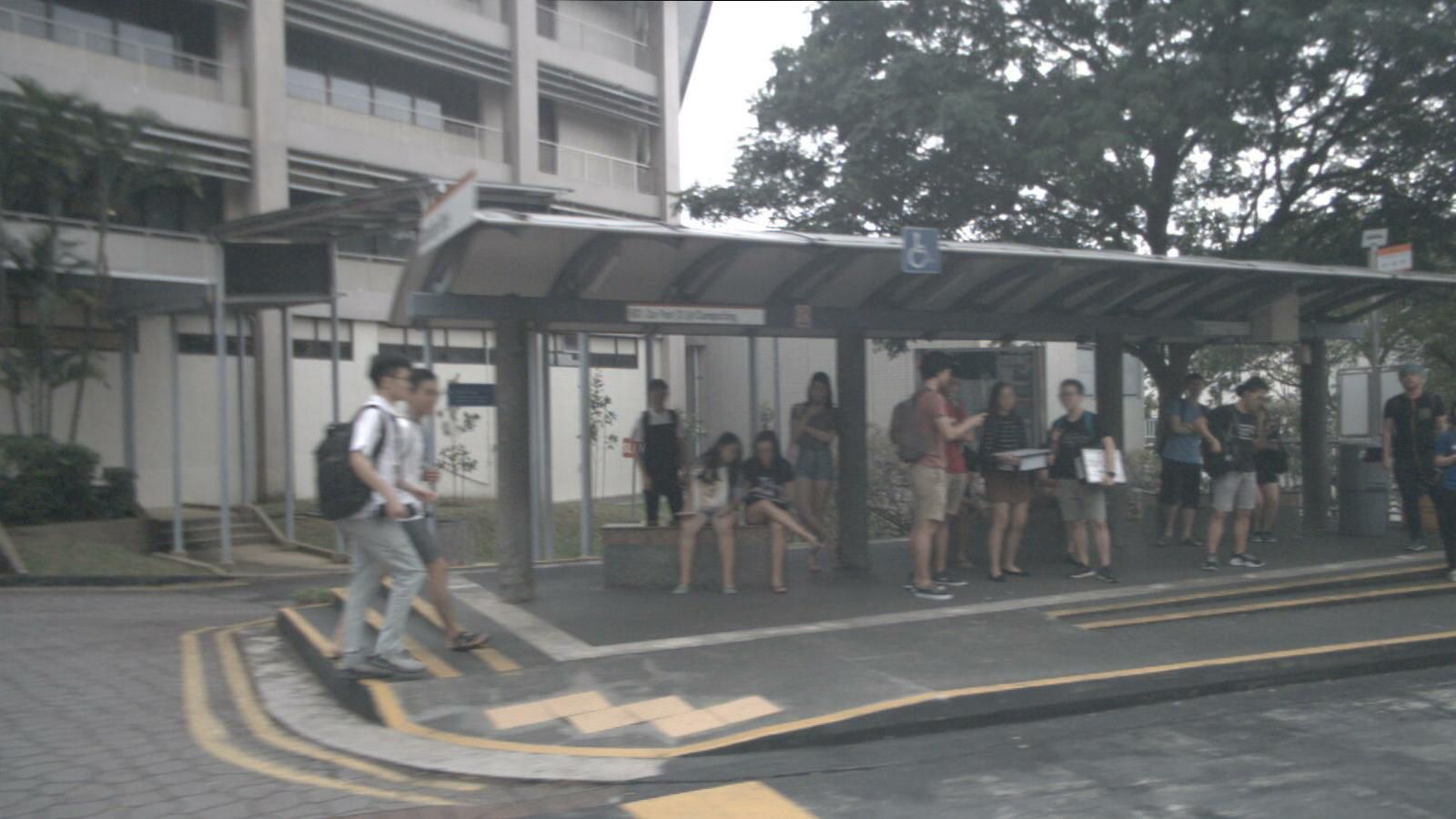}
    \includegraphics[width=\linewidth]{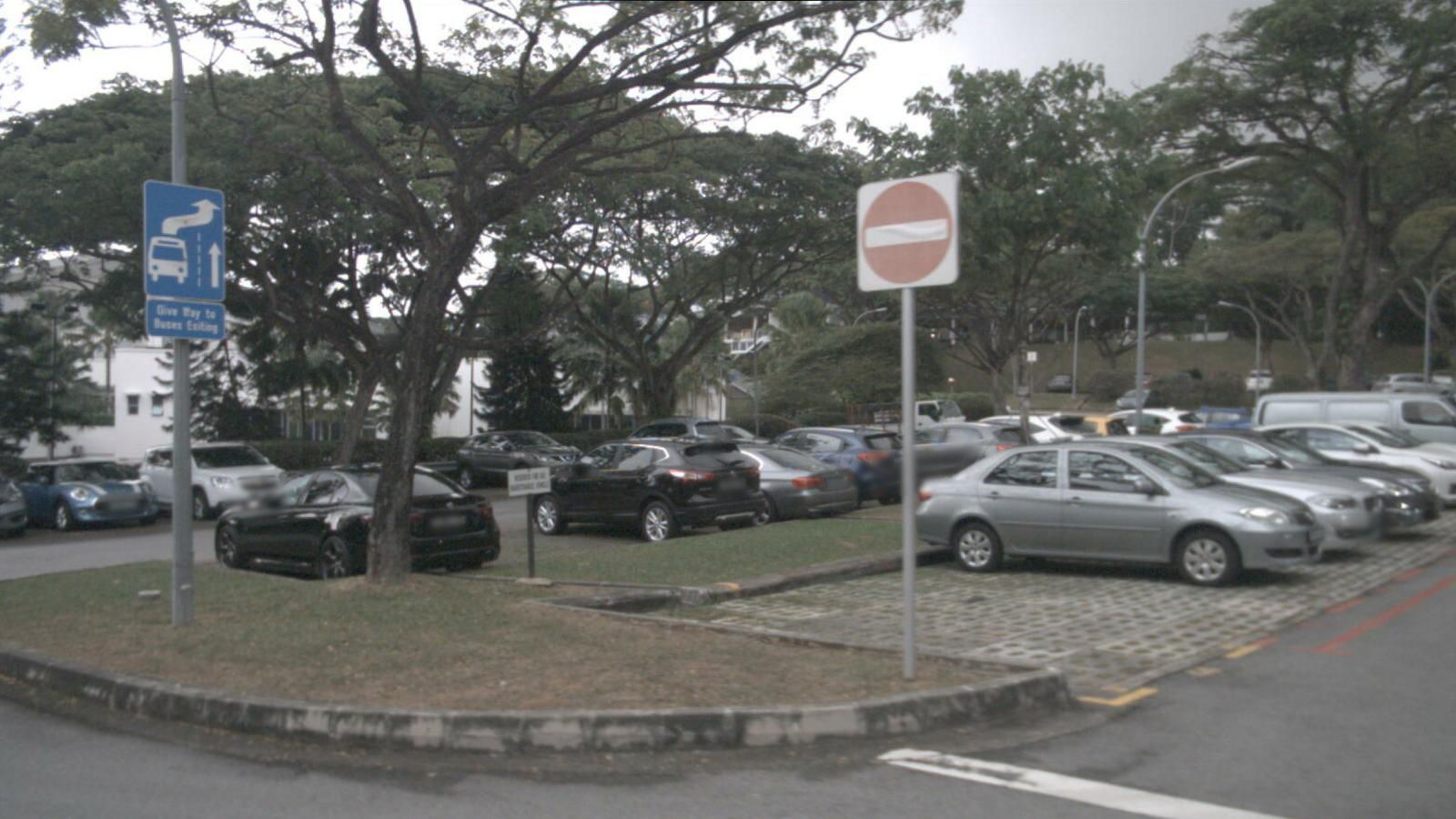}
\end{minipage}%
\hfill 
\begin{minipage}[b]{.79\textwidth}
  \centering
  \begin{subfigure}{.19\linewidth}
    \includegraphics[width=\linewidth]{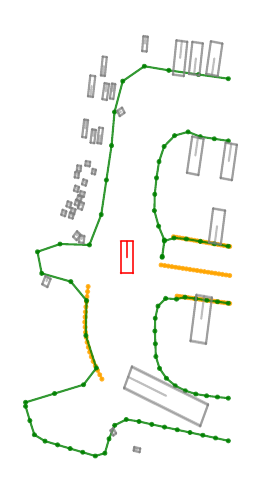}
    \caption{GT}
    \label{fig:vis2_gt}
  \end{subfigure}%
  \begin{subfigure}{.19\linewidth}
    \includegraphics[width=\linewidth]{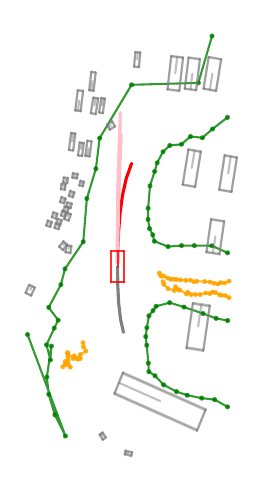}
    \caption{HiVT+MapTR}
    \label{fig:vis2_hmn}
  \end{subfigure}%
  \begin{subfigure}{.19\linewidth}
    \includegraphics[width=\linewidth]{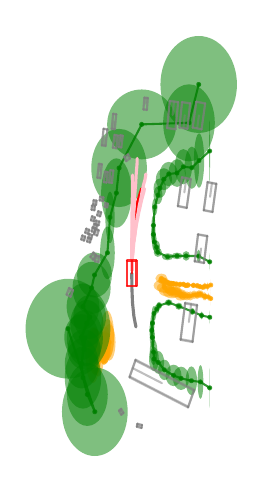}
    \caption{HiVT+MapTR}
    \label{fig:vis2_hmu}
  \end{subfigure}%
  \begin{subfigure}{.19\linewidth}
    \includegraphics[width=\linewidth]{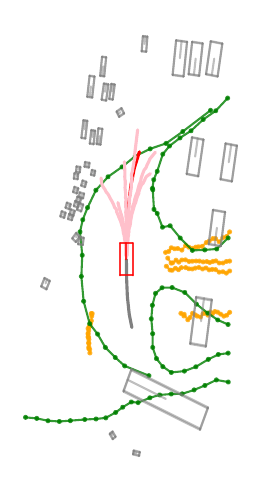}
    \caption{DenseTNT+Stream}
    \label{fig:vis2_dsn}
  \end{subfigure}
  \begin{subfigure}{.19\linewidth}
    \includegraphics[width=\linewidth]{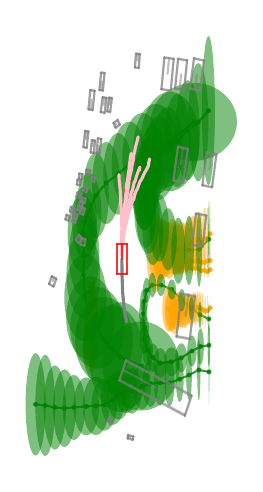}
    \caption{DenseTNT+Stream}
    \label{fig:vis2_dsu}
  \end{subfigure}
\end{minipage}
\caption{The parking lot of a bus terminal, with many occlusions from stationary vehicles. By leveraging uncertainty information, both combinations reduce overshoot, minimizing endpoint error, and tightly cluster the predicted trajectories around the GT future.}
\label{fig:vis2}


\end{figure*}

\begin{figure*}[!htbp]
\centering

\begin{minipage}[b]{.19\textwidth}
    \includegraphics[width=\linewidth]{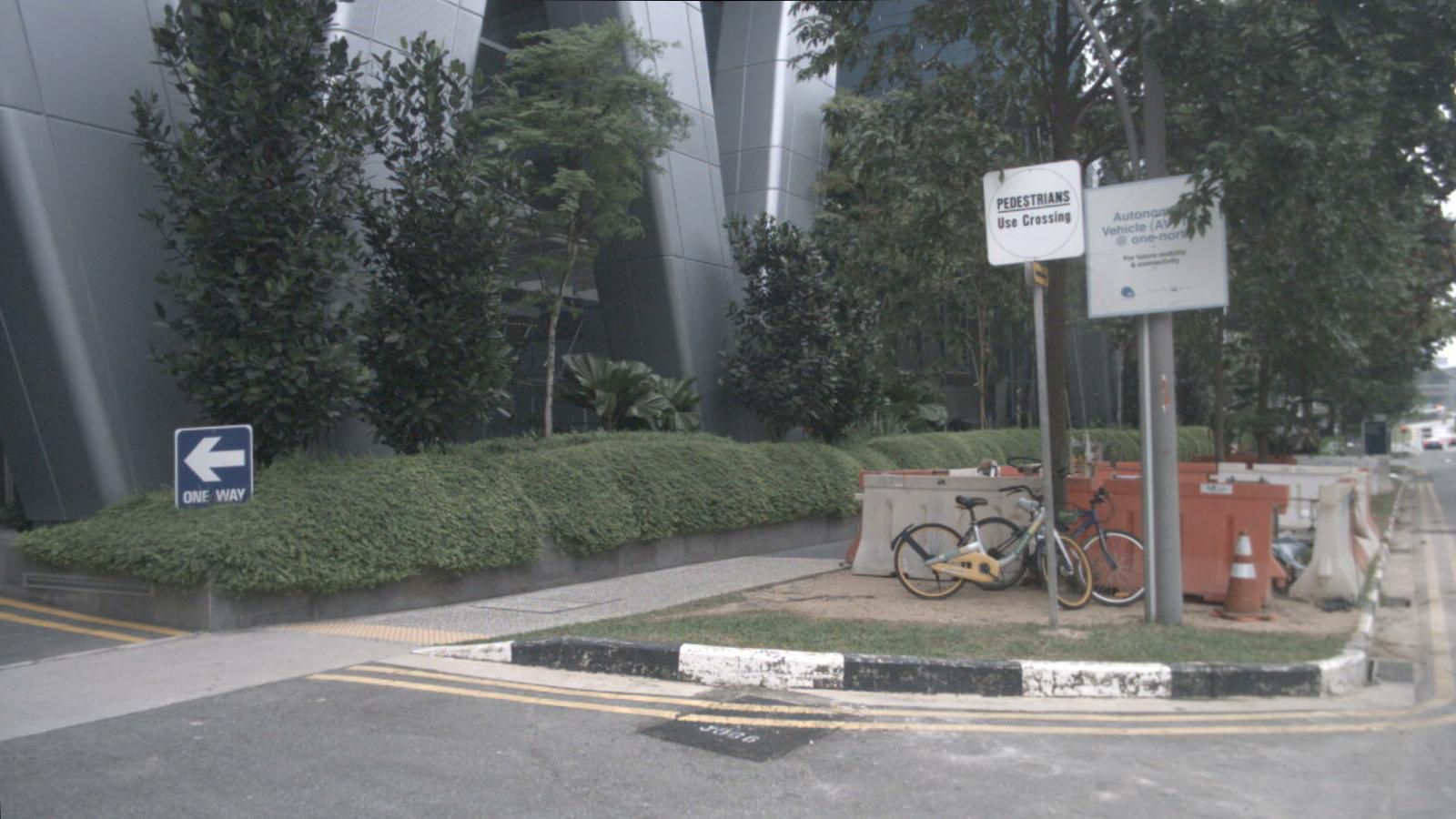}
    \includegraphics[width=\linewidth]{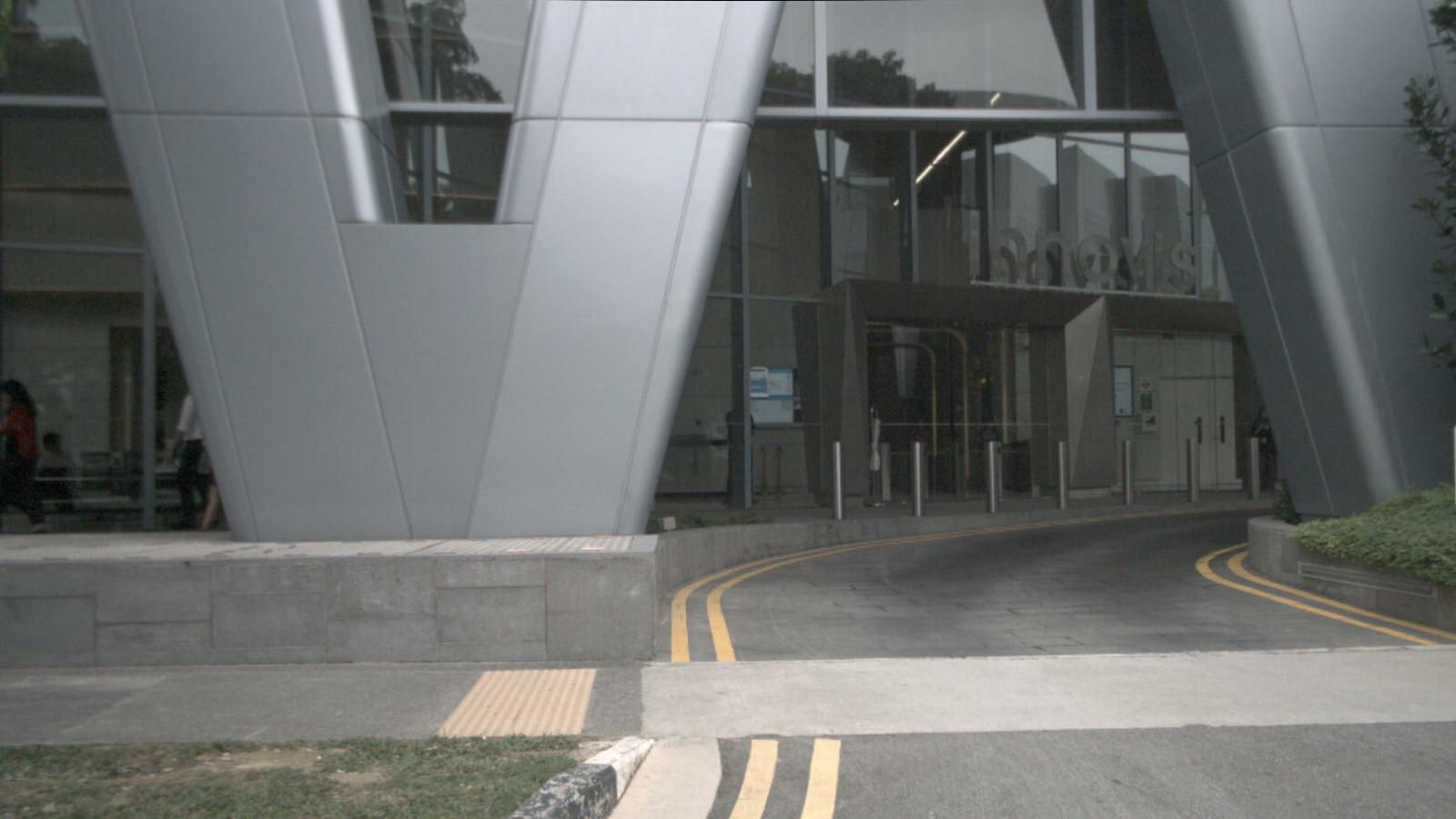}
    \includegraphics[width=\linewidth]{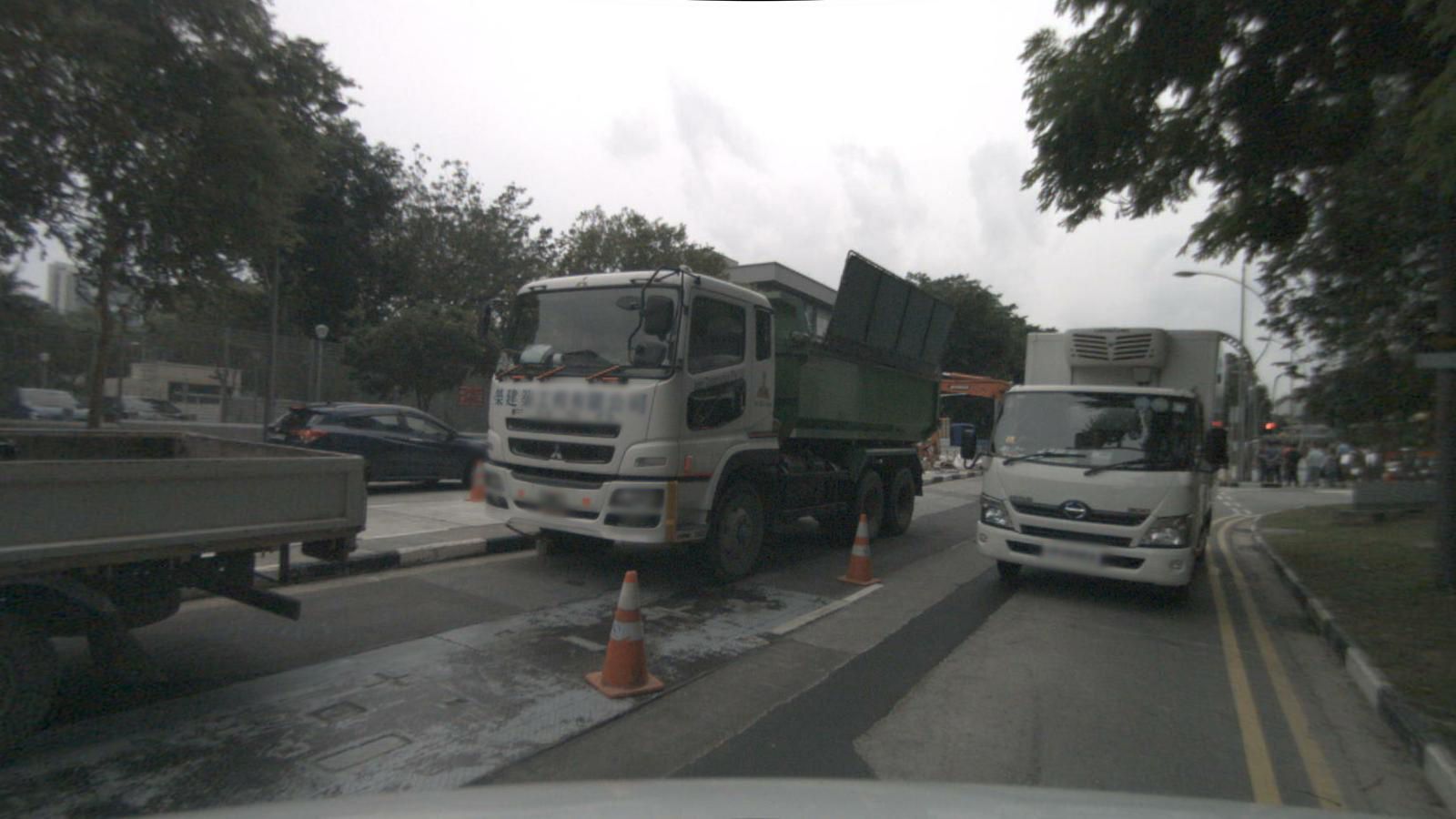}
\end{minipage}%
\hfill 
\begin{minipage}[b]{.79\textwidth}
  \centering
  \begin{subfigure}{.19\linewidth}
    \includegraphics[width=\linewidth]{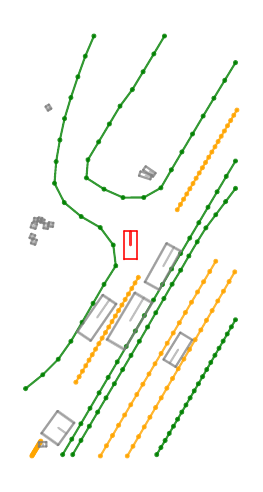}
    \caption{GT}
    \label{fig:vis3_gt}
  \end{subfigure}%
  \begin{subfigure}{.19\linewidth}
    \includegraphics[width=\linewidth]{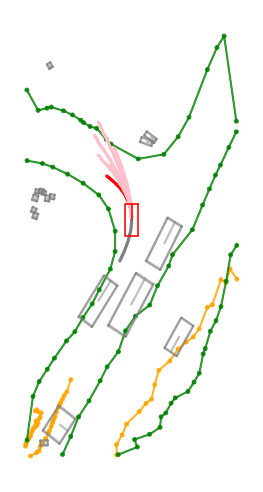}
    \caption{HiVT+MapTR}
    \label{fig:vis3_hmn}
  \end{subfigure}%
  \begin{subfigure}{.19\linewidth}
    \includegraphics[width=\linewidth]{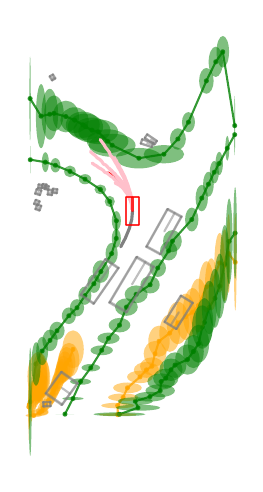}
    \caption{HIVT+MapTR}
    \label{fig:vis3_hmu}
  \end{subfigure}%
  \begin{subfigure}{.19\linewidth}
    \includegraphics[width=\linewidth]{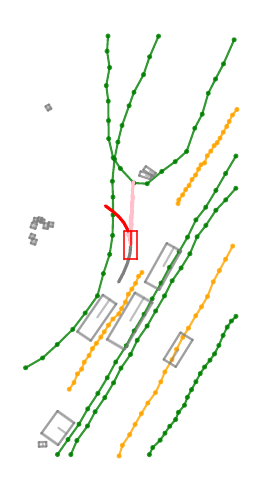}
    \caption{DenseTNT+Stream}
    \label{fig:vis3_dsn}
  \end{subfigure}
  \begin{subfigure}{.19\linewidth}
    \includegraphics[width=\linewidth]{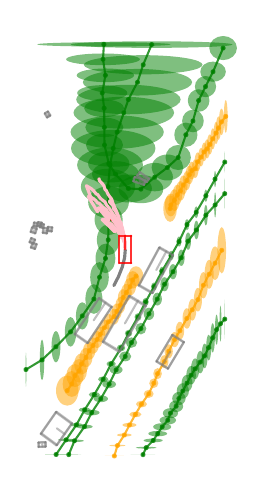}
    \caption{DenseTNT+Stream}
    \label{fig:vis3_dsu}
  \end{subfigure}
\end{minipage}
\caption{A tunnel-like entrance to a building, with significant occlusions from trucks behind the center agent. By leveraging this uncertainty information, both HiVT and DenseTNT are able to produce sensible, on-road predictions, even with significant map uncertainty.}
\label{fig:vis3}

\vspace{0.3cm}

\end{figure*}

\textbf{Prediction Visualizations.} While each model can predict trajectories for every agent, for clarity we only plot predictions for the center agent. \cref{fig:vis1} visualizes a complicated intersection with many map elements (\cref{fig:vis1_gt}). For HiVT + MapTR, predictions without map uncertainty overshoot the ground truth, directly into another lane (\cref{fig:vis1_hmn}). With map uncertainty, HiVT's predictions stay within the correct lane (\cref{fig:vis1_hmu}). For DenseTNT + StreamMapNet, predictions without map uncertainty end up offroad, ignoring the left road boundary (\cref{fig:vis1_dsn}). With map uncertainty, DenseTNT's predictions stay within the road boundary, as StreamMapNet produced them with high certainty (\cref{fig:vis1_dsu}).

\cref{fig:vis2} visualizes the parking lot of a bus terminal, containing many occlusions from stationary vehicles (\cref{fig:vis2_gt}). For HiVT + MapTR, predictions without map uncertainty significantly overshoot the ground truth, directly towards the road boundary where many motorcycles are parked (\cref{fig:vis2_hmn}). With map uncertainty, HiVT's predictions much better match the GT motion (\cref{fig:vis2_hmu}). StreamMapNet has particular difficulty mapping this environment, predicting road boundaries that pass through the middle of the road and yielding errant predictions that move towards pedestrians (\cref{fig:vis2_dsn}). With map uncertainty, DenseTNT's predictions stay within the road boundary and tightly cluster around the GT future (\cref{fig:vis2_dsu}).

\cref{fig:vis3} visualizes an interesting tunnel-like entrance to a building, with significant occlusions from trucks behind the center agent (\cref{fig:vis3_gt}). For HiVT + MapTR, predictions without map uncertainty overshoot the ground truth and collide with the road boundary (\cref{fig:vis3_hmn}). With map uncertainty, HiVT's predictions stay within the correct lane and closely match the GT future (\cref{fig:vis3_hmu}). StreamMapNet again has particular difficulty mapping this environment, predicting a road boundary that directly passes over the middle of the road, yielding errant predictions (\cref{fig:vis3_dsn}). With map uncertainty, DenseTNT's predictions nearly completely overlap with the GT future (\cref{fig:vis3_dsu}).

\section{Conclusion}
\label{sec:conclusion}

In this work, we propose a general vectorized map uncertainty formulation and extend multiple state-of-the-art online map estimation methods MapTR~\cite{MapTR}, MapTRv2~\cite{maptrv2}, and StreamMapNet~\cite{yuan2024streammapnet} to additionally output uncertainty. We systematically analyze the resulting uncertainties and find that our approach captures many sources of uncertainty (occlusion, distance to camera, time of day, and weather). Finally, we combine these online map estimation models with state-of-the-art trajectory prediction approaches (DenseTNT~\cite{GuSunEtAl2021} and HiVT~\cite{zhou2022hivt}) and show that incorporating online mapping uncertainty \emph{significantly} improves the performance and training characteristics of prediction models, by up to \textbf{50}\% and \textbf{15}\%, respectively. An exciting future research direction is leveraging these uncertainty outputs to measure the calibration of map models (similar to~\cite{IvanovicHarrisonEtAl2023}). However, this is complicated by the need for fuzzy point set matching, a challenging problem itself.

{
    \small
    \bibliographystyle{ieeenat_fullname}
    \bibliography{main}
}

\clearpage
\appendix
\setcounter{page}{1}
\maketitlesupplementary

\section{Training Details}
\label{sec:supp_training}

To account for possible differences in the rates of convergence for mapping and prediction models trained with and without uncertainty, each model's hyperparameters are tuned separately to optimize performance. The best outcomes from these individually-tuned models are compared to show the effect of integrating uncertainty. \cref{tab:supp_hyperparams} summarizes the core method hyperparameters.

In the probabilistic map estimation approaches for MapTR~\cite{MapTR} and MapTRv2~\cite{maptrv2}, we alter the loss function from the initial $\ell_1$ loss to the Negative Log-Likelihood (NLL) of the Laplacian distribution. The Laplace output is added to all layers of MapTR's Transformers. The two core reasons we chose a Laplace distribution are: it produced more accurate maps across models ($\sim$ 3-5\% better AP) and was much more numerically stable during training (\cref{fig:reb_2}).

We also adjust the regressor's loss weight from 5 to 0.03, compensating for the increased gradient norm resulting from the new loss function and difficulty in training. Given the use of a single GPU, we reduce the learning rate to 1.5E-4. Additionally, to avert gradient explosion, we clip gradients to a maximum norm of 3. All other settings are retained as per the original configurations.

For StreamMapNet~\cite{yuan2024streammapnet}, we incorporate two distinct dataset splits: the original nuScenes~\cite{CaesarBankitiEtAl2019} split and a newly proposed split. This new split is designed to address the overlapping scene challenges in the original training and validation splits~\cite{yuan2024streammapnet}. To enhance StreamMapNet's performance, we train using the new split to reduce overfitting risks and assess on the original nuScenes validation set, aligning with the scenarios used in MapTR and MapTRv2. In line with these adjustments, the loss weight of the regressor is lowered to 2, the learning rate is set to 1.25e-4, and a maximum gradient norm of 3 is maintained for clipping.

After modification, most of the map estimation models maintain their original performance. As shown in \cref{tab:map}, MapTR and MapTRv2 produce 1\% better AP when producing uncertainty. This is admittedly a small improvement, but it comes for free along with the other benefits stated in the main body of the paper.

For the HiVT~\cite{zhou2022hivt} prediction model, we have increased the dropout rate to 0.2. All other hyperparameters are unchanged. For DenseTNT~\cite{GuSunEtAl2021}, the hyperparameters are tuned separately for each combination to yield the optimal results. The hyperparameters used for different methods are shown in ~\cref{tab:hyper}.

For HiVT, we double the node dimension to account for uncertainty in both the $x$ and $y$ directions. For DenseTNT, the configurations of layer sizes and structures are maintained without significant alterations. The model utilizes a 128-dimensional vector to represent lane information, including details like vertices, intersection signals, traffic lights, etc. In our adaptation, we merely integrate uncertainty information into this raw feature vector.

\textbf{Distant Agents.} For both HiVT~\cite{zhou2022hivt} and DenseTNT~\cite{GuSunEtAl2021}, there is no special treatment for agents that are beyond map perception range. This means that some far-away agents do not have agent-lane interactions to incorporate, making the model rely only on the agent's past history, surrounding agent motion (agent-agent interactions remain unchanged), and any learned priors as a result of training with the absence of far-away map information.

\begin{table}
  \centering
  \resizebox{1\linewidth}{!}{
  \begin{tabular}{@{}l|ccc@{}}
    \toprule
    Method & Regression Loss Weight & LR & Gradient Norm \\
    \midrule
    MapTR~\cite{MapTR} & 0.03 & $1.50\text{E-}4$
 & 3  \\
    MapTRv2~\cite{maptrv2} & 0.03 & $1.50\text{E-}4$
 & 3  \\
    MapTRv2-Centerline~\cite{maptrv2} & 0.03 & $1.50\text{E-}4$ & 3  \\
    StreamMapNet~\cite{yuan2024streammapnet} & 2 & $1.25\text{E-}4$
 & 3  \\
    \bottomrule
  \end{tabular}
  }
  \vspace{-0.2cm}
  \caption{Training hyperparameters.}
  \label{tab:supp_hyperparams}

  
\end{table}

\begin{table}
  \centering
   \resizebox{0.65\linewidth}{!}{
  \begin{tabular}{@{}l|l@{}}
    \toprule
    Online HD Map Method & mAP \\
    \midrule
    MapTR~\cite{MapTR} & 0.4488  \\
    MapTR~\cite{MapTR} + Ours & 0.4525 {\color{darkpastelgreen} ($-1\%$)} \\
    \midrule
    MapTRv2~\cite{maptrv2} & 0.5540  \\ 
    MapTRv2~\cite{maptrv2} + Ours & 0.5592 {\color{darkpastelgreen} ($-1\%$)}  \\
    \midrule
    MapTRv2-Centerline~\cite{maptrv2} & 0.4789 \\
    MapTRv2-Centerline~\cite{maptrv2} + Ours & 0.4655 {\color{orange} ($+3\%$)} \\
    \midrule
    StreamMapNet~\cite{yuan2024streammapnet}  & 0.7789 \\
    StreamMapNet~\cite{yuan2024streammapnet}  + Ours & 0.7043 {\color{orange} ($+10\%$)}  \\
    \bottomrule
  \end{tabular}
    }

\vspace{-0.2cm}
  
  \caption{Map estimation performance when producing uncertainty.}
  \label{tab:map}

  
\end{table}

\begin{table}
  \centering
   \resizebox{0.85\linewidth}{!}{%
  \begin{tabular}{@{}l|l c c@{}}
    \toprule
    Online HD Map Method & LR & Batch Size & Dropout \\
    \midrule
    MapTR~\cite{MapTR} & 0.001 & 64 & 0.5 \\
    MapTR~\cite{MapTR} + Ours & 0.0005 & 64 & 0.5 \\
    \midrule
    MapTRv2~\cite{maptrv2} & 0.0005 & 64 & 0.5  \\ 
    MapTRv2~\cite{maptrv2} + Ours & 0.0005 & 64 & 0.5  \\
    \midrule
    MapTRv2-Centerline~\cite{maptrv2} & 0.001 & 64 & 0.5 \\
    MapTRv2-Centerline~\cite{maptrv2} + Ours & 0.00018&64&0.5  \\
    \midrule
    StreamMapNet~\cite{yuan2024streammapnet}  & 0.0005 & 16 & 0.1 \\
    StreamMapNet~\cite{yuan2024streammapnet}  + Ours & 0.001 & 64 & 0.5   \\
    \bottomrule
  \end{tabular}
    }

\vspace{-0.2cm}
  
  \caption{Hyperparameters chosen for different mapping methods for DenseTNT ~\cite{GuSunEtAl2021}}
  \label{tab:hyper}

  
\end{table}

\section{Additional Visualizations}
\label{sec:supp_vis}

\begin{figure*}[t]
\centering

\begin{minipage}[b]{\textwidth}
  \centering
  \begin{subfigure}{.33\linewidth}
    \includegraphics[width=\linewidth]{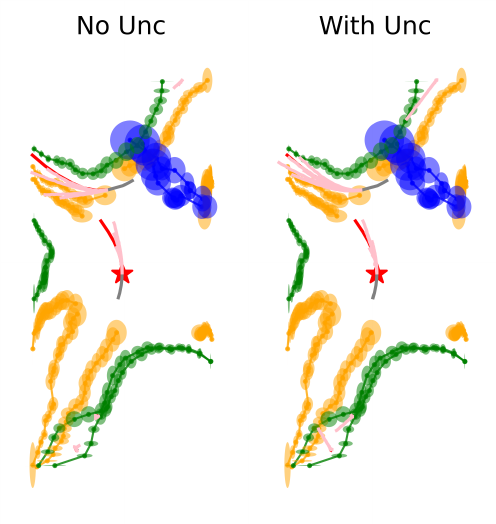}
  \end{subfigure}%
  \begin{subfigure}{.33\linewidth}
    \includegraphics[width=\linewidth]{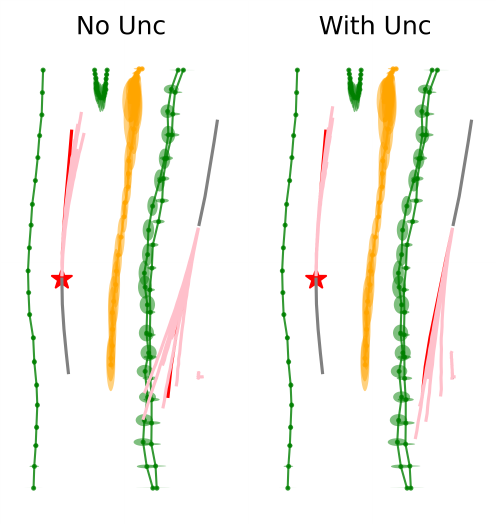}
  \end{subfigure}%
  \begin{subfigure}{.33\linewidth}
    \includegraphics[width=\linewidth]{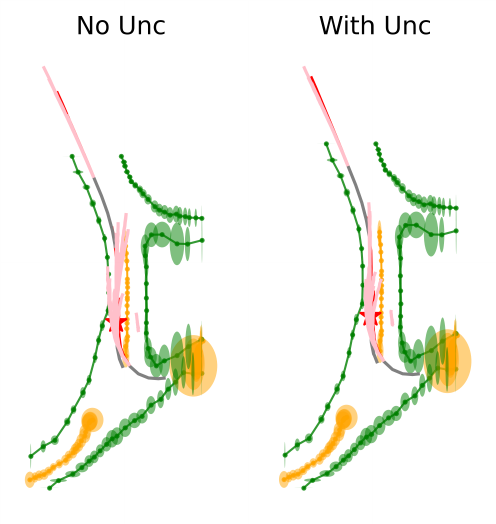}
  \end{subfigure}%
\end{minipage}
\caption{Multi-agent visualizations. \textcolor{red}{Red} indicates the GT and \textcolor{pink}{pink} shows future agent predictions. In all three scenarios, our approach produces sensible predictions for both ego and non-ego agents.}
\label{fig:reb_1}
\end{figure*}

\cref{fig:reb_1} visualizes the predictions of other agents when using multi-agent prediction models such as HiVT. 

\begin{figure*}[t]
\centering

\begin{minipage}[b]{.19\textwidth}
    \includegraphics[width=\linewidth]{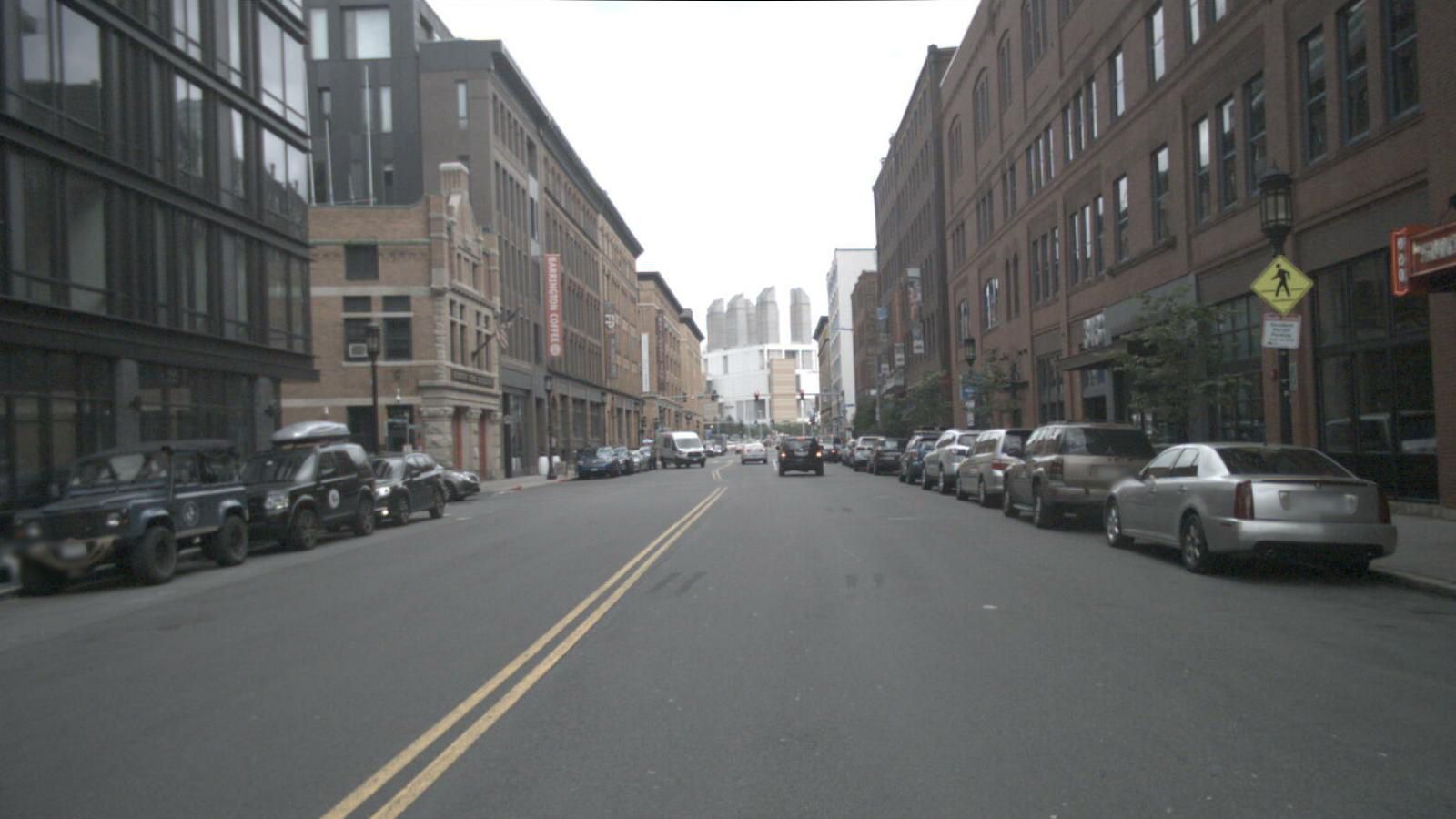}
    \includegraphics[width=\linewidth]{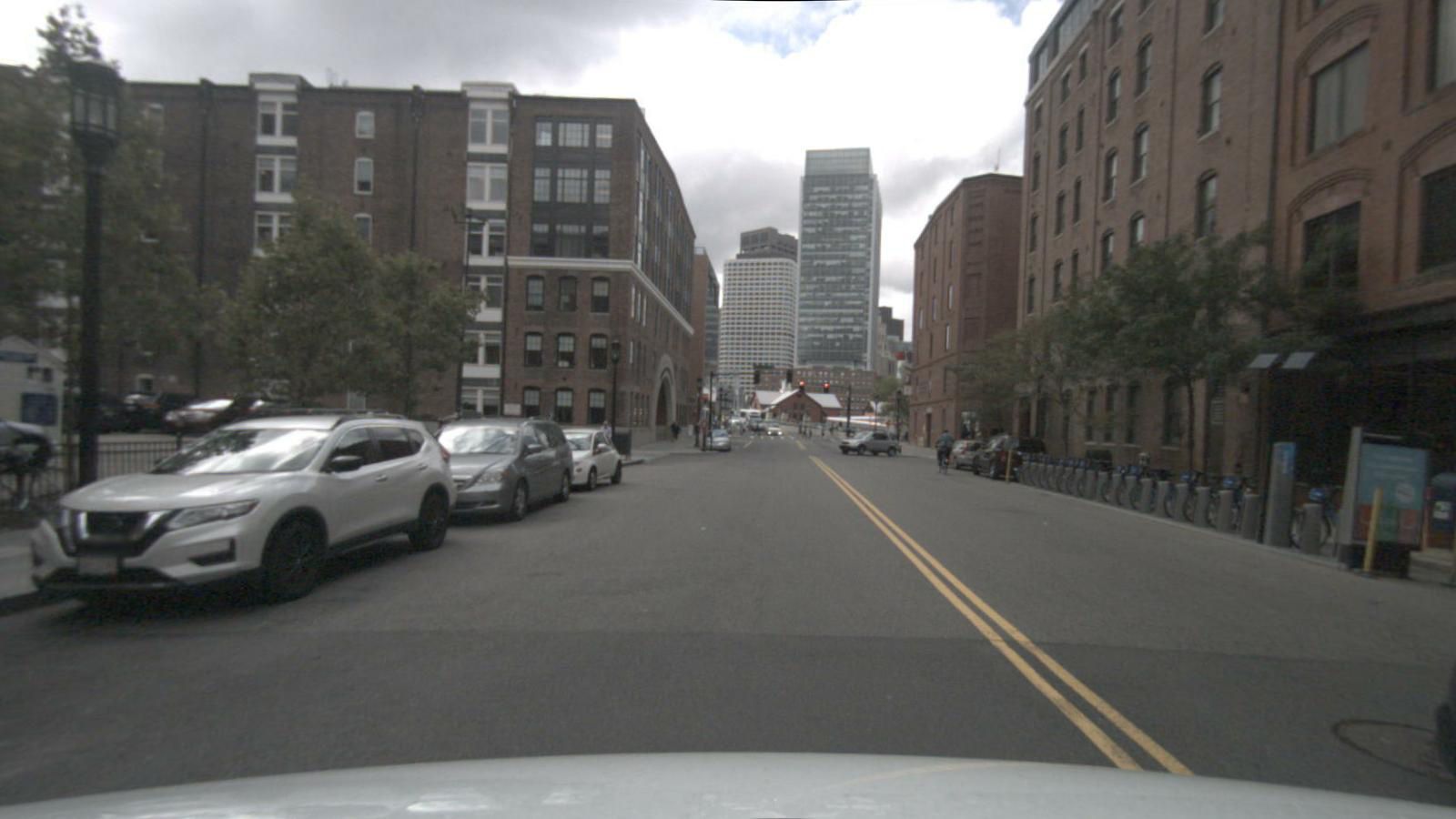}
    \includegraphics[width=\linewidth]{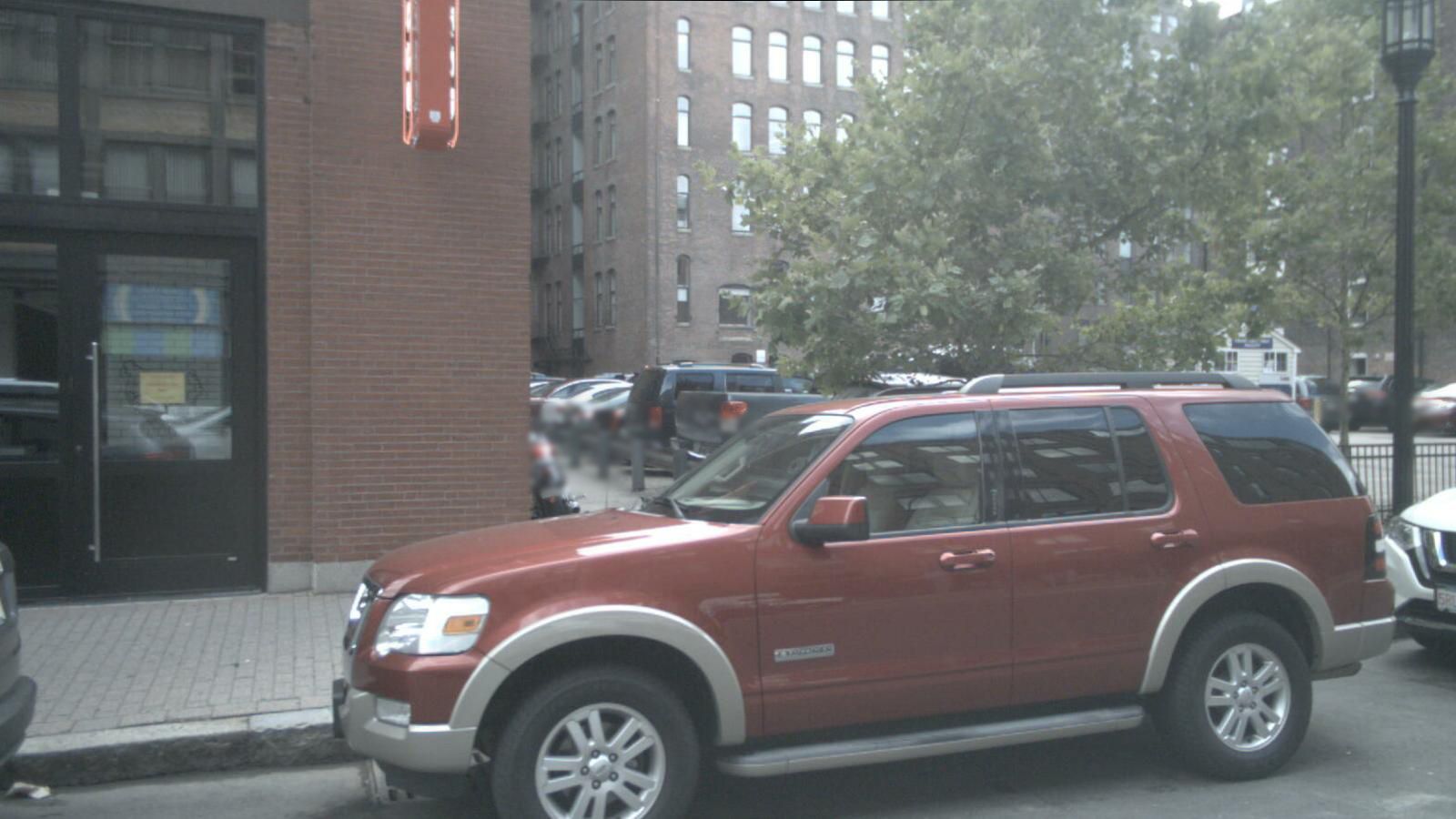}
\end{minipage}%
\hfill 
\begin{minipage}[b]{.74\textwidth}
  \centering
  \begin{subfigure}{.19\linewidth}
    \includegraphics[width=\linewidth]{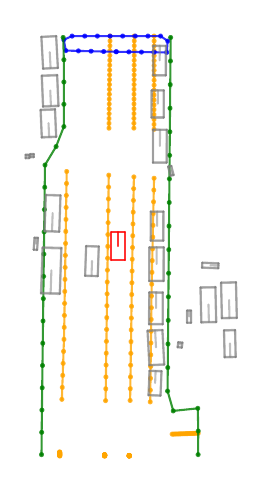}
    \caption{GT}
  \end{subfigure}%
  \begin{subfigure}{.19\linewidth}
    \includegraphics[width=\linewidth]{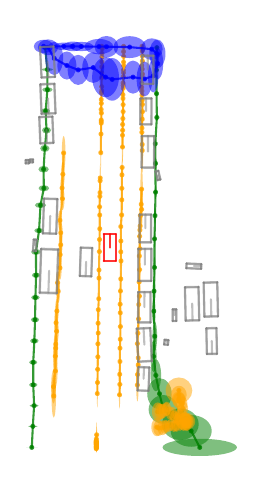}
    \caption{MapTR}
  \end{subfigure}%
  \begin{subfigure}{.19\linewidth}
    \includegraphics[width=\linewidth]{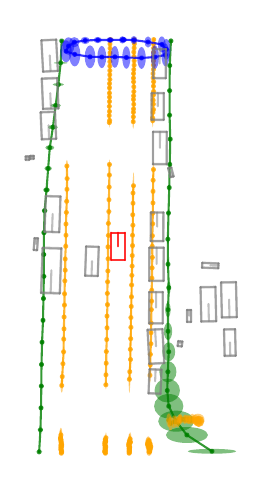}
    \caption{StreamMapNet}
  \end{subfigure}%
  \begin{subfigure}{.19\linewidth}
    \includegraphics[width=\linewidth]{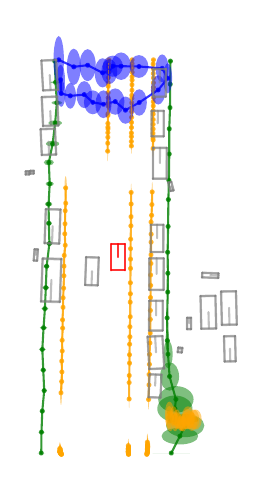}
    \caption{MapTRv2}
  \end{subfigure}
  \begin{subfigure}{.19\linewidth}
    \includegraphics[width=\linewidth]{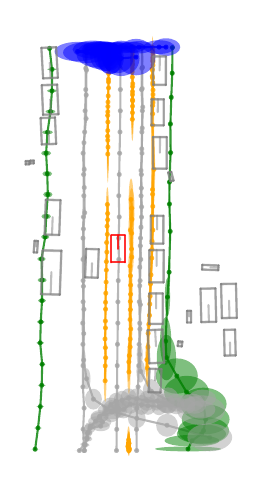}
    \caption{MapTRv2-Center}
  \end{subfigure}
\end{minipage}
\caption{A normal straight-driving scenario. Note that the parked cars on the rear right induce a larger uncertainty compared to the rear left, showing the effect of occlusions in online mapping uncertainty.}
\label{fig:supp_occlusion}
\end{figure*}

\cref{fig:supp_occlusion} shows another qualitative example of how occlusion impacts model uncertainty.

\begin{figure*}[t]
\centering
\includegraphics[width=\textwidth]{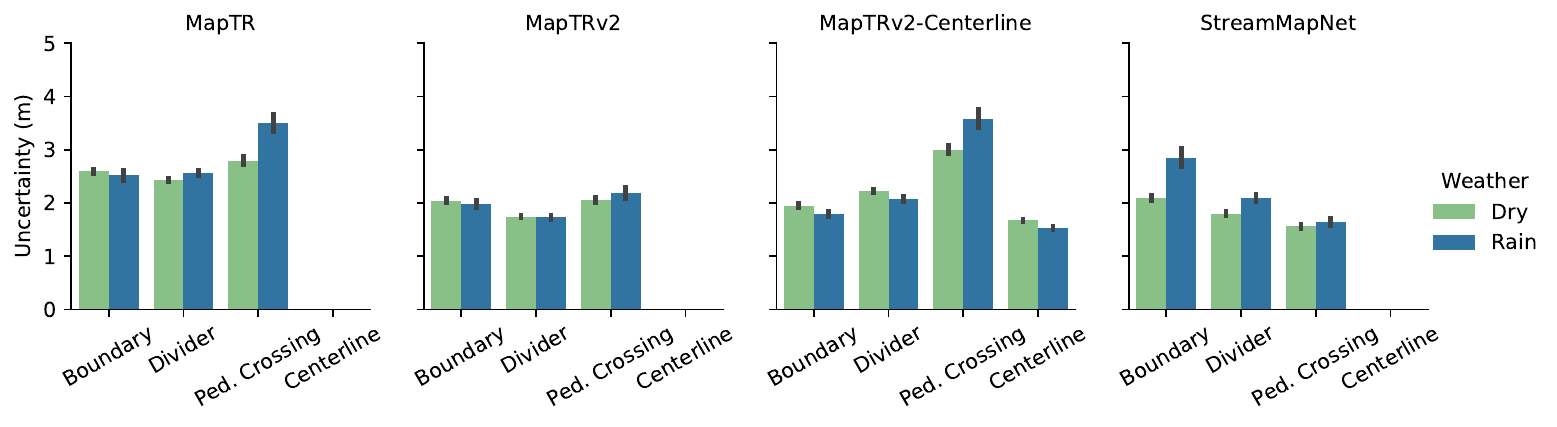}
\caption{Different weather conditions such as rain can affect the confidence with which map estimation models predict certain elements, such as pedestrian crossings for certain models. Error bars show 95\% confidence intervals.}
\label{fig:rain}
\end{figure*}

\cref{fig:rain} shows that StreamMapNet~\cite{yuan2024streammapnet} produces more uncertainty in rainy conditions, indicating potential difficulties in aggregating temporal information due to rain.

\begin{figure*}[t]
\centering
\includegraphics[width=\textwidth]{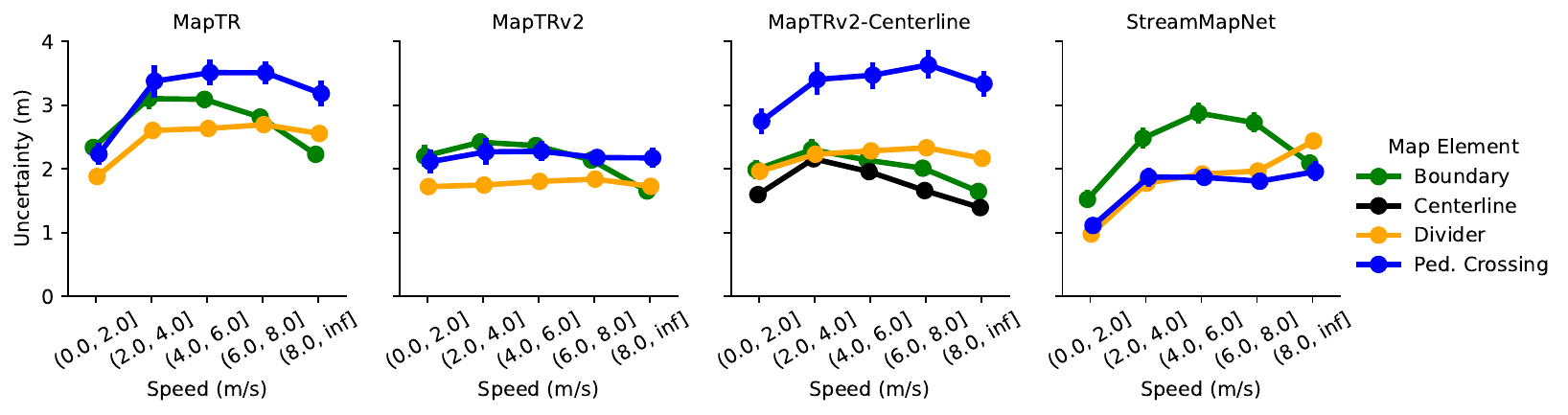}
\caption{For some scenarios, our uncertainty formulation captures the fact that uncertainty increases as the velocity of the AV increases. Error bars show 95\% confidence intervals.}
\label{fig:speed}
\end{figure*}

\cref{fig:speed} shows that current models do not have any particular lack of confidence across different AV driving speeds. 

\begin{figure*}[t]
\centering

\begin{minipage}[b]{\textwidth}
  \centering
  \begin{subfigure}{.33\linewidth}
    \includegraphics[width=\linewidth]{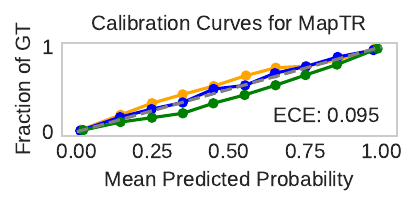}
  \end{subfigure}%
  \begin{subfigure}{.33\linewidth}
    \includegraphics[width=\linewidth]{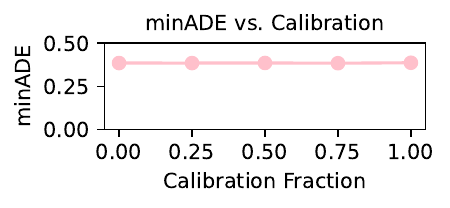}
  \end{subfigure}%
  \begin{subfigure}{.33\linewidth}
    \includegraphics[width=\linewidth]{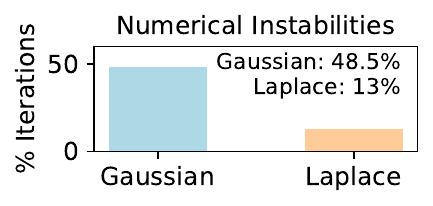}
  \end{subfigure}%
\end{minipage}
\caption{\textbf{Left:} MapTR's lane type estimates are well-calibrated for \textcolor{color1}{divider}, \textcolor{color2}{ped crossing} and \textcolor{color3}{boundary}. \textbf{Middle:} HiVT is robust to lane type miscalibration. \textbf{Right:} Laplace outputs are much more stable than Guassian to train with.}
\label{fig:reb_2}
\end{figure*}

\textbf{Calibration.} As seen in ~\cref{fig:reb_2}, MapTR's lane type estimates are well-calibrated. Further, \cref{fig:reb_2} shows that prediction methods like HiVT are robust to lane type miscalibration (evaluated by linearly interpolating between MapTR's well-calibrated probabilities and a uniform distribution over type, and predicting trajectories with the resulting probabilites). One possible hypothesis is that HiVT focuses more on the presence of lanes rather than their types.

\end{document}